\PassOptionsToPackage{table}{xcolor}
\PassOptionsToPackage{hyperfootnotes=false}{hyperref}
\documentclass{youtu}
\usepackage{mathpazo}
\usepackage[utf8]{inputenc}
\usepackage[T1]{fontenc}
\usepackage{natbib}
\hypersetup{
  pdftitle={Latent Multimodal Memory Reasoning},
  pdfauthor={Cai Ke, Xinghao Chen, Yuanchen Bei, Yuni Lai, Keyu Chen, Ruizhi Qiao, Siyu An, Junnan Dong, Ruifeng Xu, Xing Sun}
}

\usepackage{amsmath,amsfonts,bm}

\def\eqref#1{equation~\ref{#1}}

\def\1{\bm{1}}

\DeclareMathAlphabet{\mathsfit}{\encodingdefault}{\sfdefault}{m}{sl}
\SetMathAlphabet{\mathsfit}{bold}{\encodingdefault}{\sfdefault}{bx}{n}

\usepackage{url}
\usepackage{booktabs}
\usepackage{xcolor}
\usepackage{colortbl}
\usepackage{makecell} 
\usepackage{multirow}
\usepackage{graphicx}
\usepackage{wrapfig}
\usepackage{subcaption}
\usepackage{float}
\newfloat{algorithm}{t}{alg}
\floatname{algorithm}{Algorithm}
\usepackage{listings}
\providecolor{UpGreen}{HTML}{1F7A45}
\providecolor{DownRed}{HTML}{B3261E}
\providecommand{\gap}[1]{\makebox[16pt][l]{\raisebox{0.35ex}{\tiny\textcolor{DownRed}{$\downarrow$#1}}}}
\providecommand{\gain}[1]{\makebox[19pt][l]{\raisebox{0.35ex}{\tiny\textcolor{UpGreen}{$\uparrow$#1}}}}
\providecommand{\costup}[1]{\makebox[15pt][l]{\raisebox{0.35ex}{\tiny\textcolor{DownRed}{$\uparrow$#1}}}}
\providecommand{\costdown}[1]{\makebox[15pt][l]{\raisebox{0.35ex}{\tiny\textcolor{UpGreen}{$\downarrow$#1}}}}
\providecommand{\dropdn}[1]{\textsuperscript{\textcolor{DownRed}{$\downarrow$#1}}}

\usepackage{hyperref}
\usepackage{dashrule} 

\definecolor{SubHeaderGray}{RGB}{242,242,242}
\definecolor{RowPurple}{RGB}{238,234,248}
\definecolor{UpGreen}{HTML}{1F7A45}
\definecolor{DownRed}{HTML}{B3261E}

\providecolor{UpGreen}{HTML}{1F7A45}
\providecolor{DownRed}{HTML}{B3261E}
\providecommand{\gap}[1]{\makebox[19pt][l]{\raisebox{0.35ex}{\tiny\textcolor{DownRed}{$\downarrow$#1}}}}
\providecommand{\gain}[1]{\makebox[19pt][l]{\raisebox{0.35ex}{\tiny\textcolor{UpGreen}{$\uparrow$#1}}}}
\providecommand{\costup}[1]{\makebox[26pt][l]{\raisebox{0.35ex}{\tiny\textcolor{DownRed}{$\uparrow$#1}}}}
\providecommand{\costdown}[1]{\makebox[26pt][l]{\raisebox{0.35ex}{\tiny\textcolor{UpGreen}{$\downarrow$#1}}}}
\providecommand{\dropdn}[1]{\textsuperscript{\textcolor{DownRed}{$\downarrow$#1}}}

\title{Latent Multimodal Memory Reasoning}

\author{\small Cai Ke\textsuperscript{$1$}, Xinghao Chen\textsuperscript{$1,2,3$}, Yuanchen Bei\textsuperscript{$4$}, Yuni Lai\textsuperscript{$5$}, Keyu Chen\textsuperscript{$1$},\\
Ruizhi Qiao\textsuperscript{$1$}, Siyu An\textsuperscript{$1$}, Junnan Dong\textsuperscript{$1\dagger$}, Ruifeng Xu\textsuperscript{$6\dagger$}, Xing Sun\textsuperscript{$1$}}
\vspace{-5mm}
\affiliation{\footnotesize
\textsuperscript{$1$}Tencent Youtu Lab\quad
\textsuperscript{$2$}Eastern Institute of Technology\quad
\textsuperscript{$3$}The Hong Kong Polytechnic University\\
\textsuperscript{$4$}University of Illinois at Urbana-Champaign\quad
\textsuperscript{$5$}Xidian University\quad
\textsuperscript{$6$}Shenzhen Loop Area Institute\\
caike@tencent.com}

\youtufinalcopy
\begin{document}

\abstract{
Long-term personalized agents are required to reason over distributed history across modalities and evolving dialogue states. While existing flat cross-modal methods compress visual evidence prematurely and struggle with temporally stale observations, they neglect the necessity to unify multimodal and cross-session evidence. Crucially, raw multimodal memories are inherently entangled and heterogeneous, making fine-grained visual grounding and cross-modal reasoning computationally prohibitive. To this end, we present latent multimodal memory reasoning, i.e., \textsc{Lmmr}, a novel neuro-symbolic framework that shifts long-term multimodal memory organization into a continuous latent space. Specifically, $(i)$ instead of converting images to text and relying on static retrieval, \textsc{Lmmr} maps both textual hidden states and visual patch features into a shared Sparse Autoencoder concept space, establishing sparse activations as common semantic addresses to dynamically construct a turn-level multimodal memory graph. $(ii)$ A graph encoder then aggregates cross-modal and temporal dependencies across the task-specific latent subgraph, injecting the graph readout as a continuous residual rollout into the generation backbone while retaining native visual tokens to preserve fine-grained details. Extensive experiments on nine Mem-Gallery tasks and twelve MemEye settings demonstrate that \textsc{Lmmr} significantly outperforms state-of-the-art baselines in capturing complex multimodal dependencies while enabling faithful and grounded responses.

}
\maketitle
{\renewcommand{\thefootnote}{}\footnotetext{$\dagger$ Corresponding authors.}}

\section{Introduction}
Long-term agent memory must support cross-session recall, temporal reasoning, and knowledge updates~\citep{maharana-etal-2024-evaluating,wu2025longmemeval}. In multimodal settings, the goal is not merely to ensure that a model has once seen an image, but to make that image usable as evidence much later. A query may depend on a small attribute in an old image, information distributed across several dialogue turns, or which of multiple visually similar states remains valid after a subsequent update. Building on dynamic agent memory~\citep{xu2025mem,kang-etal-2025-memory,chhikara2025mem0}, the model must determine whether evidence from different modalities refers to the same entity, how that evidence changes over time, and which observations still support the current answer.

Most multimodal memory systems adopt one of two visual interfaces. Some convert images into captions, semantic tags, or summaries and store them together with dialogue text~\citep{jain2025augustus,wang2025mirix}. This representation is convenient for compressing, indexing, and updating long histories, but it can erase exact colors, counts, spatial relations, and instance-level distinctions. Others retain native images or image pointers, retrieve a bounded set of historical turns using text and image embeddings, and reload the corresponding visual content at answer time~\citep{chen2022murag,yeo2026universalrag}. Recent systems extend this design with contextual user profiles, graph-structured or hybrid memories, reliability-aware retrieval, and iterative memory management~\citep{long2026seeing,lu2026mma,feng2026m2a,liu2026omni}. These mechanisms improve external memory management but leave evidence integration inside the answer model largely unresolved.

These approaches expose a tension between visual fidelity and memory organization. Semantic summaries efficiently represent stable, high-level facts but often lose fine-grained perceptual evidence. Native images preserve such evidence, yet similarity-based retrieval can return observations that are visually related but temporally stale. Recent benchmark findings~\citep{bei2026mem,guo2026memeye} reveal a consistent pattern: native visual evidence becomes more valuable as the required detail increases, but retaining and retrieving images alone remains insufficient when the answer depends on an evolving visual state. A system may recover the correct object without identifying its currently valid state. Retrieval must therefore account for both temporal validity and the reasoning needed to connect the recovered evidence across changing visual states and multiple dialogue sessions.

\begin{wrapfigure}{r}{0.59\linewidth} 
\centering
\vspace{-5mm}
\includegraphics[width=\linewidth]{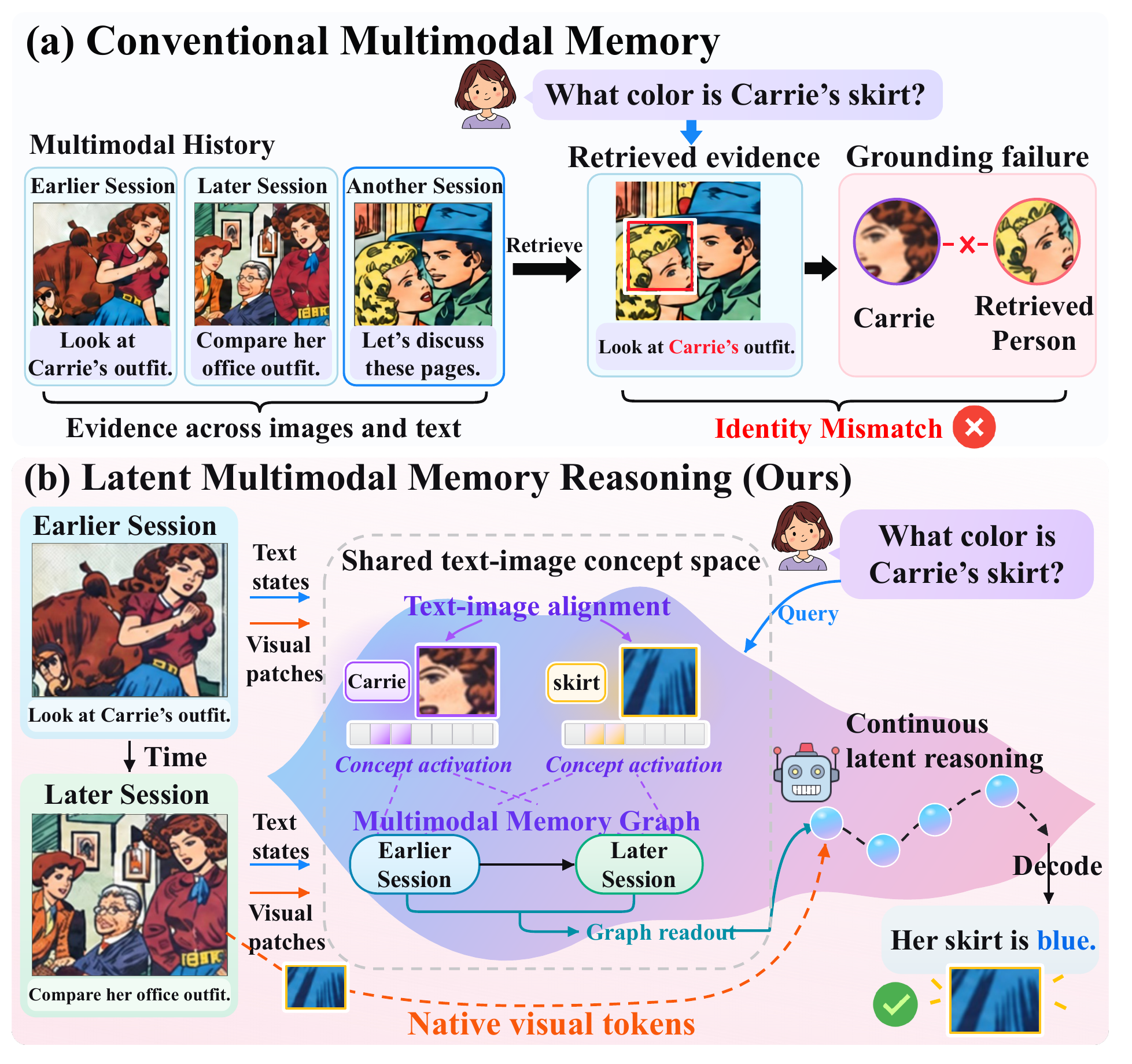}
\caption{Comparison of conventional multimodal memory and our LMMR framework. (a) Conventional retrieval may select evidence associated with the wrong entity, leading to grounding errors. (b) LMMR aligns text states and visual patches in a shared concept space, reasons over connected memories in the latent space, and preserves native visual tokens for grounded generation.}
\label{intro}
\end{wrapfigure}

We argue that a query-conditioned multimodal memory state is missing between retrieval and language. To construct such a state, we introduce Latent Multimodal Memory Reasoning (LMMR), a neuro-symbolic framework that organizes multimodal history in a continuous latent space. LMMR maps textual hidden states and visual patch features through modality-specific projectors into a shared sparse autoencoder (SAE) dictionary, using sparse activations as common semantic addresses across modalities. These addresses dynamically connect visual evidence to its dialogue turns in a query-conditioned memory graph. A graph encoder aggregates cross-modal and temporal dependencies, while retaining native visual tokens in the multimodal prefix to preserve details omitted by concept abstraction. The graph readout is injected as a residual into a continuous latent rollout, allowing the model to integrate textual, visual, and temporal evidence before a grounding token returns it to lexical generation. In short, concepts provide semantic addresses, native visual tokens retain perceptual evidence, and latent states support evidence reasoning.

We evaluate LMMR on Mem-Gallery and MemEye, covering nine long-term-memory task categories and twelve settings with varying levels of visual detail and memory complexity. LMMR consistently outperforms existing methods and strong baselines across both benchmarks. The results support organizing distributed evidence into a query-conditioned state before generation.

\textbf{The contributions of this work can be summarized as follows:}

\begin{itemize}

\item We identify the need to organize retrieved text and images into a coherent memory state that preserves cross-modal context and temporal validity for the current query.

\item We propose LMMR to connect text and images through shared SAE concepts, preserve their contextual and temporal relations in a dialogue-turn graph, and combine structured memory with native visual details during latent reasoning before producing an answer.

\item Experiments on two benchmarks show improved performance over existing methods and strong baselines, highlighting the complementary roles of preserving fine-grained visual details and reasoning over structured, temporally grounded multimodal evidence.

\end{itemize}
\section{Related Work}

\paragraph{Long-Term Agent Memory.}\looseness=-1
Memory for large language model (LLM) agents has evolved from simple retrieval and summarization toward more structured mechanisms for reflection, hierarchical storage, and memory updating. Early systems store past interactions in external memory and retrieve relevant entries for subsequent decisions, while reflection-based methods consolidate experience into reusable high-level knowledge~\citep{park2023generative,shinn2023reflexion,packer2023memgpt,zhong2024memorybank}. More recent work explicitly manages the memory lifecycle through dynamic linking, tiered storage, selective consolidation, and continual refinement~\citep{xu2025mem,kang-etal-2025-memory,chhikara2025mem0,ke2025flexibly,ke2026dynamic,liang2026meta,liu2026simplemem}. In parallel, long-horizon benchmarks have moved beyond isolated recall to evaluate memory across multiple sessions, including factual retrieval, temporal reasoning, knowledge updates, and abstention~\citep{maharana-etal-2024-evaluating,wu2025longmemeval,hu2026evaluating}. Despite this progress, most existing systems represent memory as text, summaries, or structured textual notes. This design supports efficient indexing and revision, but visual experience must usually be compressed into language before it can be stored or reused.

\paragraph{Multimodal Long-Term Memory.}
Multimodal retrieval-augmented generation (RAG) extends external memory to mixed image--text collections, allowing a query to retrieve native visual evidence together with textual context~\citep{chen2022murag,yeo2026universalrag}. Recent multimodal agent systems further introduce persistent visual observations, graph-structured memories, contextual user profiles, confidence-aware retrieval, hybrid semantic--episodic stores, and iterative memory management~\citep{jain2025augustus,wang2025mirix,bo2025agentic,long2026seeing,lu2026mma,feng2026m2a,liu2026omni,xie2026fluxmem}. Related directions include modality-aware integration for knowledge-based visual question answering~\citep{dong2024modality} and native multimodal modeling~\citep{an2026toward}. These systems improve access to visual history, but they still face a basic trade-off. Converting images into semantic records makes long histories easier to organize and update, but may discard instance-level or pixel-level details; retaining native images preserves these details, but similarity-based retrieval can return visually related yet stale or conflicting evidence. Earlier multimodal dialogue benchmarks primarily evaluate localized multi-turn or multi-image understanding~\citep{liu2024mmdu,xue2025mmrc}, whereas Mem-Gallery extends evaluation to multi-session multimodal memory extraction, reasoning, and knowledge management~\citep{bei2026mem}. MemEye further isolates whether systems preserve indispensable visual details and recover the valid state from evolving visual histories~\citep{guo2026memeye}. Together, these studies show that effective multimodal memory requires more than storing or retrieving images. \textbf{Unlike prior systems that treat retrieved text and images as independent context, LMMR formulates multimodal memory as a query-conditioned latent state: shared concepts link evidence across modalities, while a memory graph preserves dialogue dependencies and temporal validity.}

\paragraph{Graph and Latent Reasoning.}
Complementary work studies hierarchy-aware graph question answering~\citep{dong2023hierarchy}, knowledge-graph prompting~\citep{dong2024knowgpt}, cost-efficient knowledge-based question answering~\citep{dong2024cost}, and unified GraphRAG agents~\citep{dong2026youtu}. Latent chain-of-thought research examines reasoning beyond language~\citep{chen2026reasoninglanguagecomprehensivesurvey} and effective supervision through an information-theoretic lens~\citep{chen2026what}. LMMR focuses on connecting multimodal memories and reasoning over their temporal dependencies in a latent state.
\section{Problem Formulation}
\label{sec:problem}

We study question answering over multimodal long-term interaction histories. Let
$\mathcal{H}=\{r_1,r_2,\ldots,r_M\}$ denote a multi-session dialogue history, where
$r_i=(x_i,\mathcal{I}_i,\tau_i)$ contains the textual interaction $x_i$, a set of images
$\mathcal{I}_i$, and its timestamp $\tau_i$. Let $q$ denote the current query, which may
also contain a visual observation, and let $y=(y_1,\ldots,y_N)$ be the target response.
We use $f_\theta$ to denote the backbone multimodal language model and $E(\cdot)$ its
token embedding function.

\noindent\fbox{\parbox{\dimexpr\linewidth-2\fboxsep-2\fboxrule\relax}{
Given a potentially long multimodal history $\mathcal{H}$ and a query $q$, the objective
is to generate a response $y$ grounded in the relevant historical evidence. Rather than
directly decoding from retrieved memory items, the model first constructs a
query-conditioned memory state $\mathbf{M}_q$ that preserves visual details, dialogue
context, and temporal relations. At inference time, $\mathbf{M}_q$ is constructed only
from the accessible history and the current query, with no access to annotated clues,
gold image identifiers, or information derived from ground-truth answers.
}}

Formally, $G_\phi$ produces $\mathbf{M}q$ from the history $\mathcal{H}{\leq q}$ truncated at the question anchor, and the backbone generates an answer conditioned on this query-specific memory state:

\begin{equation}
\mathbf{M}_q = G_\phi(\mathcal{H}_{\leq q},q),
\qquad
p_{\theta,\phi}(y\mid q,\mathcal{H}_{\leq q})
=
\prod_{n=1}^{N}
p_{\theta,\phi}
\left(
y_n\mid y_{<n},q,\mathbf{M}_q
\right).
\label{eq:problem}
\end{equation}

\section{Method}
\label{sec:method}

\begin{figure*}[!t]
\centering
\includegraphics[width=\textwidth]{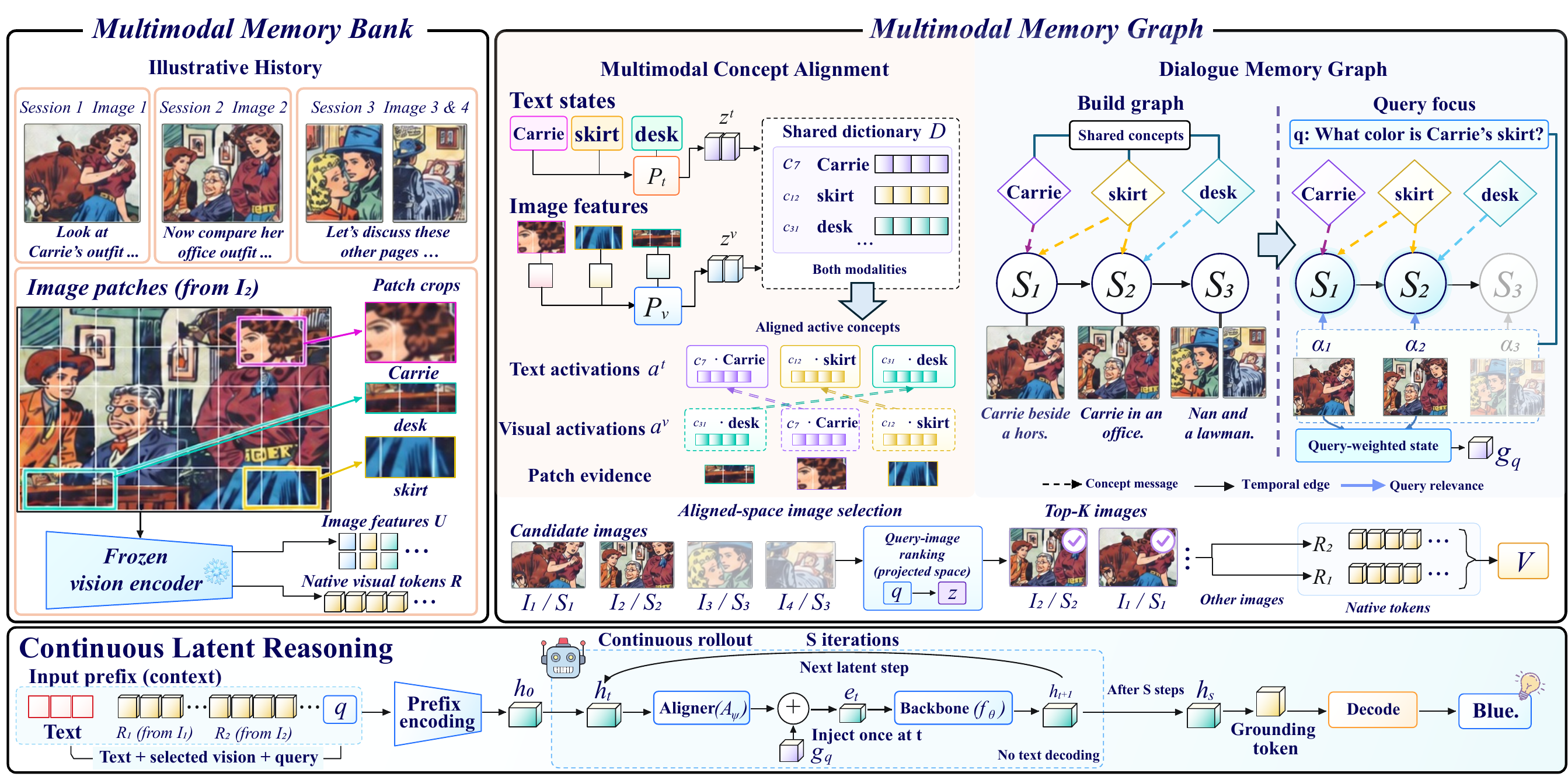}
\caption{Overview of LMMR for reasoning over long-term multimodal memories.}
\label{method}
\end{figure*}

\subsection{Overview}

LMMR is a neuro-symbolic framework that organizes multimodal history in a continuous latent space before answer generation. A memory bank retains textual context, visual patch features, native visual tokens, and temporal provenance. Modality-specific projectors map text states and visual patches into a shared SAE dictionary, whose sparse activations serve as semantic addresses for a query-conditioned, turn-level memory graph. A graph encoder aggregates cross-modal and temporal dependencies over this task-specific subgraph. Its readout enters the generation backbone as a residual in a continuous latent rollout, while native visual tokens preserve fine-grained evidence. Concepts provide semantic addresses, the graph preserves dialogue context, and latent states integrate the retrieved evidence before grounded answer generation.

\subsection{Multimodal Memory Construction}
\label{sec:memory}

A long interaction history may contain many images, only a few of which are relevant to
the current query. Passing every image to the backbone is expensive and introduces
irrelevant visual content, whereas replacing all images with captions can remove
fine-grained evidence. We therefore preserve two complementary representations of each
historical image.

For an image $I_{ij}$ appearing in turn $r_i$, a frozen visual encoder $E_v$ produces a
compact patch representation $\mathbf{U}_{ij}$ and a sequence of native visual tokens
$\mathbf{R}_{ij}$:

\begin{equation}
\mathbf{U}_{ij},\mathbf{R}_{ij}=E_v(I_{ij}),
\qquad
\mathbf{U}_{ij}\in\mathbb{R}^{P\times d_v},
\quad
\mathbf{R}_{ij}\in\mathbb{R}^{L_{ij}\times d}.
\label{eq:visual_memory}
\end{equation}

The compact representation is used for memory indexing and concept modeling, while the
native tokens retain the visual details consumed by the multimodal backbone. Each memory
entry stores
$m_{ij}=(c_{ij},\mathbf{U}_{ij},\mathbf{R}_{ij},\mathrm{id}_{ij},i,\tau_i)$,
where $c_{ij}$ is the image caption and $(i,\tau_i)$ records its dialogue provenance.
Textual turns are stored together with the same round and temporal information.

For each query, we first construct a bounded candidate bank $\mathcal{B}_q$ using a
lightweight combination of query relevance and recency. This stage reduces the search
space but does not expose every native visual token to the backbone. The following SAE
module selects a small set of query-relevant images from $\mathcal{B}_q$; only their
native visual tokens are inserted into the multimodal prefix. This two-level design
keeps long histories tractable without reducing all visual evidence to text.

\subsection{Shared SAE Alignment and Dialogue Memory Graph}
\label{sec:sae_graph}

Textual hidden states and visual patch features follow different distributions, making
their raw similarities unreliable for cross-modal memory selection. Moreover, treating
images as isolated entries discards the dialogue context and temporal state in which
they appeared. We address both problems by learning common concept coordinates and
organizing aligned evidence by dialogue turn and temporal context rather than indexing each image as an isolated memory.

\paragraph{Shared concept coordinates.}
Let $\mathbf{u}^{t}$ denote a textual state and $\mathbf{u}^{v}$ a pooled visual feature. Modality-specific projectors $P_t$ and $P_v$ map them into a common space, after which both modalities are encoded using the same learnable SAE dictionary $\mathbf{D}\in\mathbb{R}^{C\times d_c}$:

\begin{equation}
\mathbf{z}^{m}=P_m(\mathbf{u}^{m}),\qquad \mathbf{a}^{m}=\operatorname{TopKSoftmax}\left(\mathbf{D}\overline{\mathbf{z}}^{m}/\tau_s,k\right),\qquad m\in\{t,v\}.
\label{eq:sae_encoding}
\end{equation}

Here, $\overline{\mathbf{z}}$ denotes $\ell_2$ normalization and $\mathbf{a}^{m}$ contains at most $k$ nonzero activations. Text and vision use different projectors and reconstruction heads but share the same dictionary, allowing them to retain modality-specific information while referring to common concept coordinates. We optimize the SAE with $\mathcal{L}_{\mathrm{SAE}}=\mathcal{L}_{\mathrm{rec}}+\lambda_{\mathrm{align}}\mathcal{L}_{\mathrm{NCE}}$, where $\mathcal{L}_{\mathrm{rec}}$ reconstructs each modality in its projected space and $\mathcal{L}_{\mathrm{NCE}}$ aligns paired image--caption codes through noise-contrastive estimation.

The query and candidate images are compared in the aligned space using $s_j=\cos(\mathbf{z}_q^t,\mathbf{z}_j^v)+\lambda_{\mathrm{lex}}s_j^{\mathrm{lex}}$, where $s_j^{\mathrm{lex}}$ is an optional caption-based lexical prior. The top-$K$ images form the selected set $\mathcal{V}_q$.

\paragraph{Dialogue-turn graph.}
An image is meaningful not only through its visual content but also through the dialogue event in which it appeared. We therefore attach each selected image to its source turn. For turn $i$, contextualized textual states are pooled into $\mathbf{z}_i^t$, while the aligned representations of its associated images are aggregated into $\mathbf{z}_i^v$. Using layer normalization (LN), their gated combination forms the initial source-turn node $\mathbf{x}_i^{(0)}=\operatorname{LN}(\mathbf{z}_i^t+\sigma(g_v)\mathbf{z}_i^v)$.

At each graph layer, a node receives a concept message $\mathbf{m}_i^{C}$ from its nearest SAE dictionary entries and a temporal message $\mathbf{m}_i^{T}$ from the current and earlier dialogue turns:

\begin{equation}
\mathbf{x}_i^{(\ell+1)}=\operatorname{LN}\left(\mathbf{x}_i^{(\ell)}+\mathbf{m}_i^{C,(\ell)}+\mathbf{m}_i^{T,(\ell)}\right).
\label{eq:graph_update}
\end{equation}

After $L$ layers, the query assigns each node a relevance weight $\alpha_i=\operatorname{softmax}_i(\cos(\mathbf{z}_q^t,\mathbf{x}_i^{(L)})/\tau_q)$. The final graph readout is

\begin{equation}
\mathbf{g}_q=W_g\left(\mathbf{z}_q^t+\sum_i\alpha_i\mathbf{x}_i^{(L)}\right).
\label{eq:graph_readout}
\end{equation}

The concept message connects evidence expressed in different modalities, whereas the temporal message distinguishes earlier observations from later updates. The resulting readout $\mathbf{g}_q$ encodes both query relevance and the temporal context needed to interpret the selected multimodal evidence.

\subsection{Latent Multimodal Memory Reasoning}
\label{sec:latent}

Retrieving relevant evidence does not guarantee that it will be used correctly. Under direct decoding, the model must integrate multimodal evidence and verbalize the answer simultaneously. LMMR instead introduces a short continuous reasoning stage before generating text.

The native visual tokens of images in $\mathcal{V}_q$ are inserted into the image placeholders of their source turns. Together with the selected dialogue text, image captions, and current query, they form the multimodal prefix $\mathcal{P}_q$. The backbone jointly encodes this prefix as $(\mathbf{H}^{P},\mathrm{KV}_0)=f_\theta(\mathcal{P}_q)$, where $\mathrm{KV}_0$ is the key--value cache. Its final state initializes the latent rollout as $\mathbf{h}_0=\mathbf{H}^{P}_{\mathrm{last}}$.

At each latent step, an aligner $A_\psi$ maps the current hidden state back to the input-embedding space. At a fixed step $t^\star$, the graph readout supplies an additive residual to the aligned input:

\begin{equation}
\mathbf{e}_t=A_\psi(\mathbf{h}_t)+\mathbf{1}[t=t^\star]\mathbf{g}_q,\qquad (\mathbf{h}_{t+1},\mathrm{KV}_{t+1})=f_\theta(\mathbf{e}_t;\mathrm{KV}_t).
\label{eq:latent_rollout}
\end{equation}

Residual injection lets structured memory influence the reasoning trajectory without introducing additional textual tokens. After $S$ latent steps, LMMR appends a regular grounding token and resumes autoregressive decoding. The native visual tokens provide perceptual detail, the dialogue graph provides structure, and the latent rollout provides computation before the answer is expressed.

\subsection{Training Objective}
\label{sec:objective}

The visual encoder remains frozen, while the modality projectors, shared SAE dictionary, dialogue graph, and latent aligner are jointly optimized. The language backbone is trained using low-rank adaptation (LoRA), with answer supervision and auxiliary losses combined as

\begin{equation}
\mathcal{L}=\mathcal{L}_{\mathrm{ans}}+\lambda_{\mathrm{SAE}}\mathcal{L}_{\mathrm{SAE}}+\lambda_{\mathrm{graph}}\left(1-\cos(\mathbf{g}_q,\mathbf{h}_{t^\star})\right).
\label{eq:total_loss}
\end{equation}

Here, $\mathcal{L}_{\mathrm{ans}}$ is the token-level negative log-likelihood of the target answer. The SAE objective establishes common cross-modal coordinates, while graph consistency encourages the structured readout to align with the corresponding backbone state at the designated graph-injection step.

\raggedbottom
\section{Experiments}

\subsection{Experimental Setup}
\label{sec:setup}

\noindent\textbf{Datasets and benchmarks.} We evaluate LMMR on two multimodal long-term memory benchmarks: Mem-Gallery~\citep{bei2026mem}, spanning nine long-term memory task categories at two backbone scales, and MemEye~\citep{guo2026memeye}, spanning twelve settings that range from scene-level to pixel-level visual evidence and from atomic retrieval to evolutionary synthesis. Appendices~\ref{app:benchmarks} and~\ref{app:metrics} provide dataset statistics, evaluation protocols, and future-information leakage controls.

\noindent\textbf{Comparison systems.} On Mem-Gallery we reuse the thirteen systems benchmarked in~\citep{bei2026mem}: Full Context (Text) (FC (T)), First-In-First-Out (FIFO), NaiveRAG, Generative Agents (Gen.Agent )~\citep{park2023generative}, Reflexion~\citep{shinn2023reflexion}, MemGPT~\citep{packer2023memgpt}, A-Mem~\citep{xu2025mem}, MemoryOS (MemOS)~\citep{kang-etal-2025-memory}, Full Context (Multimodal) (FC (V)), MuRAG~\citep{chen2022murag}, UniversalRAG~\citep{yeo2026universalrag}, Neural Graph Memory (NGM)~\citep{fisher2025neural} and AUGUSTUS~\citep{jain2025augustus}. MemEye baselines include: FC, Semantic RAG (SRAG), Reflexion, Generative Agents, MemOS, A-Mem, SimpleMem (SM)~\citep{liu2026simplemem} with textual (T) and multimodal (V) variants, Multimodal Memory Agent (MMA)~\citep{lu2026mma}, and the dual-layer Multimodal Memory Agent (M2A)~\citep{feng2026m2a}.

\noindent\textbf{Evaluation metrics.} We report token-level F1 (the harmonic mean of precision and recall), unigram Bilingual Evaluation Understudy (BLEU-1), exact match (EM), and LLM-as-a-Judge (\mbox{L-Judge}) scores. Mem-Gallery uses Qwen2.5-72B-Instruct as its judge; MemEye uses GPT-5.2 for open-ended answers and averages multiple-choice EM over four answer rotations. These metrics capture lexical overlap, exact recall, and semantic correctness; Appendices~\ref{app:metrics} and~\ref{app:prompt-judge} provide complete scoring protocols and judge prompts for both benchmarks.

\noindent\textbf{Implementation.} LMMR is instantiated on Qwen2.5-VL-3B/7B~\citep{bai2025qwen2} for Mem-Gallery and on Qwen3-VL-8B~\citep{bai2025qwen3} for MemEye. The visual encoder stays frozen, the backbone is adapted with LoRA~\citep{hu2021lora},
and the projectors, the shared dictionary, the dialogue-turn graph, and the latent aligner are trained jointly.
Appendix~\ref{app:setup} lists the hyperparameters. Mem-Gallery uses five-fold cross-fitting over conversation scenarios; MemEye is evaluated zero-shot using a model trained on Mem-Gallery, without MemEye training data.

\subsection{Main Results}
\label{sec:main_results}

\paragraph{Larger backbones benefit more from latent memory reasoning.}
The widening margin over MuRAG from 3B to 7B in Table~\ref{tab:memgallery_overall_size_comparison} suggests that LMMR makes better use of additional model capacity. A larger backbone can more effectively integrate the supplied multimodal evidence before decoding, allowing latent reasoning to contribute beyond evidence retrieval alone. This contrasts with the modest gains of most retrieval baselines and the degradation of full-context variants: scaling the reader is most useful when its computation can turn historical evidence into an answer.

\begin{figure}[!t]
  \centering
  \begin{subfigure}[b]{0.49\textwidth}
    \centering
    \includegraphics[width=\linewidth]{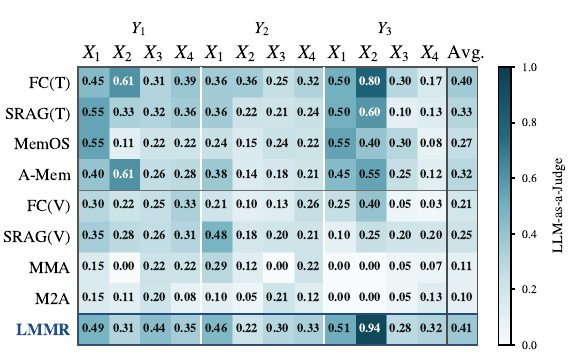}
    \caption{Open-ended L-Judge.}
    \label{fig:memeye-judge}
  \end{subfigure}\hfill
  \begin{subfigure}[b]{0.49\textwidth}
    \centering
    \includegraphics[width=\linewidth]{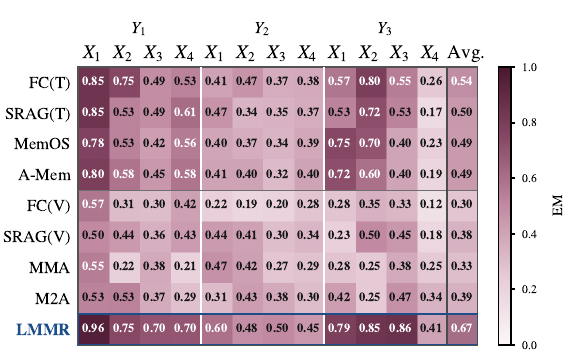}
    \caption{Multiple-choice exact match.}
    \label{fig:memeye-em}
  \end{subfigure}
  \caption{MemEye results with Qwen3-VL-8B: (a)~open-ended L-Judge; (b)~multiple-choice EM. Baseline results taken from~\citep{guo2026memeye}. Appendix~\ref{app:memeye-full} gives all per-setting results.}
  \label{fig:memeye}
\end{figure}

\paragraph{Transfer exposes a gap between recognizing and generating an answer.}
The stronger improvement on multiple-choice questions in Fig.~\ref{fig:memeye} suggests that LMMR transfers evidence useful for distinguishing candidate answers, while recovering a precise answer without candidates remains harder. Caption-based memory remains competitive when coarse descriptions suffice, but cannot consistently preserve the details needed for visual discrimination. The remaining errors motivate better grounding during generation, beyond recognizing evidence among candidate answers.

\begin{table*}[!t]
\centering
\caption{\textbf{Mem-Gallery results across backbone scales.} Best and second-best scores are \textbf{bold} and \underline{underlined}. Baseline results taken from \citep{bei2026mem}. Appendix~\ref{app:memgallery-full} reports per-task results.}
\label{tab:memgallery_overall_size_comparison}

\vspace{2pt}
\scriptsize
\setlength{\tabcolsep}{0.07cm}
\renewcommand{\arraystretch}{1.12}

\resizebox{\linewidth}{!}{%
\begin{tabular*}{\linewidth}{
@{}
>{\raggedright\arraybackslash}p{0.26\linewidth}
*{2}{
*{3}{>{\centering\arraybackslash}p{\dimexpr(\linewidth-0.26\linewidth-24\tabcolsep)/12\relax}}
>{\centering\arraybackslash}p{\dimexpr(\linewidth-0.26\linewidth-24\tabcolsep)/12+4pt\relax}
>{\centering\arraybackslash}p{\dimexpr(\linewidth-0.26\linewidth-24\tabcolsep)/12-4pt\relax}
>{\centering\arraybackslash}p{\dimexpr(\linewidth-0.26\linewidth-24\tabcolsep)/12\relax}
}
@{}
}
\toprule

\multirow{2}{*}{\textbf{Method}}
& \multicolumn{6}{c}{\textbf{Qwen-2.5-VL-3B}}
& \multicolumn{6}{c}{\textbf{Qwen-2.5-VL-7B}} \\

\cmidrule(lr){2-7}
\cmidrule(lr){8-13}

& \textbf{F1}
& \textbf{BLEU-1}
& \textbf{EM}
& \textbf{L-Judge}
& \textbf{Avg}
& \textbf{$\Delta$}
& \textbf{F1}
& \textbf{BLEU-1}
& \textbf{EM}
& \textbf{L-Judge}
& \textbf{Avg}
& \textbf{$\Delta$} \\

\midrule

\rowcolor{SubHeaderGray}
\multicolumn{13}{@{}l}{
\textit{\textbf{Text-based Memory Methods}}
} \\

\rowcolor{white}
\quad FC (T)
& 37.98 & 34.92 & 26.24 & 45.41 & 36.14
& \textcolor{UpGreen}{$\uparrow$30.03}
& 36.25 & 32.79 & 25.19 & 43.31 & 34.39
& \textcolor{UpGreen}{$\uparrow$41.79} \\

\quad FIFO
& 29.86 & 26.89 & 20.40 & 35.21 & 28.09
& \textcolor{UpGreen}{$\uparrow$38.08}
& 27.24 & 24.08 & 18.35 & 33.69 & 25.84
& \textcolor{UpGreen}{$\uparrow$50.34} \\

\quad NaiveRAG
& 58.32 & 53.33 & 39.33 & 68.56 & 54.89
& \textcolor{UpGreen}{$\uparrow$11.28}
& 59.74 & 54.41 & 41.61 & 72.41 & 57.04
& \textcolor{UpGreen}{$\uparrow$19.14} \\

\multicolumn{13}{@{}c@{}}{\vspace{-3.5pt}\hdashrule[1.3ex]{\linewidth}{0.35pt}{2.5pt 2.5pt}} \\

\quad Gen.Agent~\citep{park2023generative}
& 40.13 & 36.20 & 26.71 & 45.97 & 37.25
& \textcolor{UpGreen}{$\uparrow$28.92}
& 38.25 & 34.22 & 26.13 & 46.43 & 36.26
& \textcolor{UpGreen}{$\uparrow$39.92} \\

\quad Reflexion~\citep{shinn2023reflexion}
& 37.93 & 34.87 & 26.18 & 45.38 & 36.09
& \textcolor{UpGreen}{$\uparrow$30.08}
& 36.19 & 32.79 & 25.07 & 43.07 & 34.28
& \textcolor{UpGreen}{$\uparrow$41.90} \\

\quad MemGPT~\citep{packer2023memgpt}
& 41.65 & 37.94 & 27.00 & 53.83 & 40.11
& \textcolor{UpGreen}{$\uparrow$26.06}
& 52.82 & 47.92 & 34.02 & 73.06 & 51.96
& \textcolor{UpGreen}{$\uparrow$24.22} \\

\quad A-Mem~\citep{xu2025mem}
& 60.59 & 55.05 & 43.03 & 68.86 & 56.88
& \textcolor{UpGreen}{$\uparrow$9.29}
& 62.28 & 56.29 & 42.96 & 74.31 & 58.96
& \textcolor{UpGreen}{$\uparrow$17.22} \\

\quad MemOS~\citep{kang-etal-2025-memory}
& 60.13 & 54.62 & 41.03 & 71.62 & 56.85
& \textcolor{UpGreen}{$\uparrow$9.32}
& 61.09 & 55.34 & 42.78 & 76.13 & 58.84
& \textcolor{UpGreen}{$\uparrow$17.34} \\

\midrule

\rowcolor{SubHeaderGray}
\multicolumn{13}{@{}l}{
\textit{\textbf{Multimodal Memory Methods}}
} \\

\rowcolor{white}
\quad FC (V)
& 36.65 & 33.60 & 24.96 & 42.72 & 34.48
& \textcolor{UpGreen}{$\uparrow$31.69}
& 33.54 & 29.99 & 22.79 & 41.29 & 31.90
& \textcolor{UpGreen}{$\uparrow$44.28} \\
\multicolumn{13}{@{}c@{}}{\vspace{-3.5pt}\hdashrule[1.3ex]{\linewidth}{0.35pt}{2.5pt 2.5pt}} \\
\quad MuRAG~\citep{chen2022murag}
& \underline{66.79}
& \underline{61.94}
& \underline{47.28}
& \underline{77.56}
& \underline{63.39}
& \textcolor{UpGreen}{$\uparrow$2.78}
& \underline{69.66}
& \underline{64.32}
& \underline{49.85}
& \underline{82.29}
& \underline{66.53}
& \textcolor{UpGreen}{$\uparrow$9.65} \\

\quad NGM~\citep{fisher2025neural}
& 64.43 & 59.53 & 46.70 & 74.63 & 61.32
& \textcolor{UpGreen}{$\uparrow$4.85}
& 66.91 & 62.00 & 49.44 & 78.61 & 64.24
& \textcolor{UpGreen}{$\uparrow$11.94} \\

\quad AUGUSTUS~\citep{jain2025augustus}
& 63.83 & 58.81 & 45.41 & 75.28 & 60.83
& \textcolor{UpGreen}{$\uparrow$5.34}
& 66.10 & 60.69 & 47.57 & 77.97 & 63.08
& \textcolor{UpGreen}{$\uparrow$13.10} \\

\quad UniversalRAG~\citep{yeo2026universalrag}
& 64.79 & 59.95 & 45.94 & 76.62 & 61.83
& \textcolor{UpGreen}{$\uparrow$4.34}
& 68.27 & 62.86 & 49.27 & 80.16 & 65.14
& \textcolor{UpGreen}{$\uparrow$11.04} \\

\rowcolor{RowPurple}
\quad \textbf{LMMR (Ours)}
& \textbf{70.17}
& \textbf{65.21}
& \textbf{51.72}
& \textbf{77.56}
& \textbf{66.17}
& \textbf{--}
& \textbf{79.10}
& \textbf{74.79}
& \textbf{63.47}
& \textbf{87.35}
& \textbf{76.18}
& \textbf{--} \\

\bottomrule
\end{tabular*}}
\end{table*}

\providecolor{UpGreen}{HTML}{1F7A45}
\providecolor{DownRed}{HTML}{B3261E}
\providecommand{\dropdn}[1]{\textsuperscript{\textcolor{DownRed}{$\downarrow$#1}}}

\providecommand{\gap}[1]{\makebox[19pt][l]{\raisebox{0.35ex}{\tiny\textcolor{DownRed}{$\downarrow$#1}}}}
\providecommand{\gain}[1]{\makebox[19pt][l]{\raisebox{0.35ex}{\tiny\textcolor{UpGreen}{$\uparrow$#1}}}}

\subsection{Ablation Study}
\label{sec:ablation}

\begin{wraptable}{r}{0.5\linewidth}
\vspace{-\intextsep}
    \centering
    \scriptsize
    \renewcommand{\arraystretch}{1.15}
    \caption{Ablation study of LMMR on Mem-Gallery using Qwen2.5-VL-7B.}
    \label{tab:lmmr_ablation}
    \setlength{\tabcolsep}{0.4mm}{
    \resizebox{\linewidth}{!}{%
\begin{tabular}{@{}lcccc@{}}
        \toprule
        \textbf{Variant} & \textbf{F1} & \textbf{EM} & \textbf{BLEU-1} & \textbf{L-Judge} \\
        \midrule
        \rowcolor{RowPurple}
        \textbf{LMMR (Full)} & \textbf{79.10} & \textbf{63.47} & \textbf{74.79} & \textbf{87.35} \\
        \midrule
        w/o Text Memory & 46.27\dropdn{32.83} & 32.79\dropdn{30.68}
        & 43.16\dropdn{31.63} & 50.82\dropdn{36.53} \\
        w/o Visual Memory & 72.71\dropdn{6.39} & 55.82\dropdn{7.65}
        & 68.46\dropdn{6.33} & 79.63\dropdn{7.72} \\
        w/o Latent Reasoning & 59.17\dropdn{19.93} & 38.92\dropdn{24.55}
        & 53.61\dropdn{21.18} & 78.81\dropdn{8.54} \\
        \bottomrule
    \end{tabular}}}
\end{wraptable}

\paragraph{Evidence access and evidence integration play complementary roles.}
The ablations in Table~\ref{tab:lmmr_ablation} suggest that dialogue text supplies the context in which visual observations become useful, while native images preserve details absent from that context. Direct decoding still falls short despite receiving the memory prefix, separating evidence access from evidence use. The complementary effects support integrating dialogue context and visual details before verbalization, so that retrieval supplies evidence and latent computation integrates it before answering. Removing latent reasoning reduces F1 by 19.93 points, compared with 6.39 points without visual memory, highlighting the importance of integrating the available evidence.

\subsection{Analysis}
\label{sec:analysis}

\begin{figure}[!t]
  \centering
  \begin{subfigure}[t]{0.365\textwidth}
    \centering
    \includegraphics[width=\linewidth]{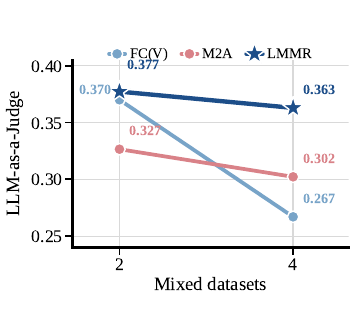}
    \caption{Mixed-memory scaling.}
    \label{fig:memeye_scaling}
  \end{subfigure}\hfill
  \begin{subfigure}[t]{0.603\textwidth}
    \centering
    \includegraphics[width=\linewidth]{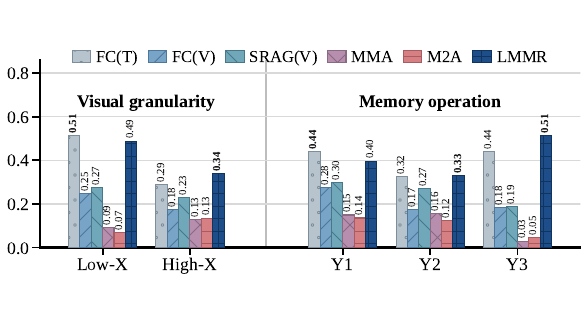}
    \caption{Region-level L-Judge performance.}
    \label{fig:memeye_region}
  \end{subfigure}
  \caption{MemEye open-ended experiments with the Qwen3-VL-8B backbone.}
  \label{fig:memeye_lmmr}
\end{figure}

\paragraph{Selective memory access limits cross-topic interference.}
In Fig.~\ref{fig:memeye_scaling}, stability under topic mixing is consistent with query-conditioned selection reducing interference from unrelated memories. The contrast with full visual context suggests that selective access preserves useful evidence as histories become longer and more diverse, beyond merely reducing the input length.

\paragraph{Visual detail matters most when memories must be reconciled.}
The granularity and reasoning breakdown in Fig.~\ref{fig:memeye_region} links LMMR's advantage to evidence that coarse summaries cannot fully express and observations whose validity changes over time. Captions can describe a scene yet omit the attribute that distinguishes an updated state from an earlier one. Retaining native visual evidence while reasoning across dialogue turns addresses both requirements, explaining why the benefit is more pronounced for fine-grained evidence and evolutionary synthesis than for coarse recognition.

\begin{figure}[!t]
\centering
\begin{minipage}[t]{0.48\linewidth}
\vspace{0pt}\centering
\includegraphics[width=\linewidth]{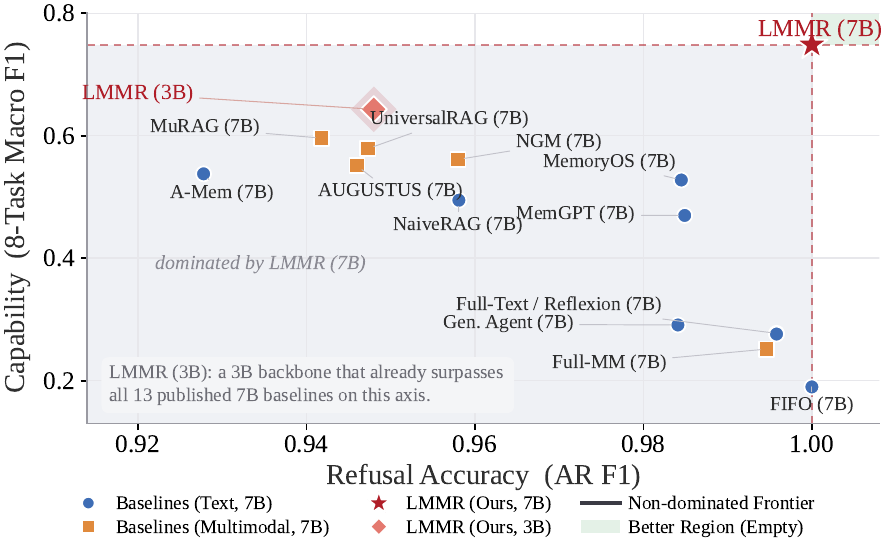}
\caption{Capability and answer refusal on Mem-Gallery with Qwen2.5-VL-7B.}
\label{fig:memgallery_pareto}
\end{minipage}\hfill
\begin{minipage}[t]{0.5\linewidth}
\vspace{0pt}\centering
\captionsetup{type=table}
\centering
    \scriptsize
    \setlength{\tabcolsep}{1.5pt}
    \renewcommand{\arraystretch}{1.12}
    \caption{Cost per query on Mem-Gallery.}
    \label{tab:efficiency}
    \resizebox{\linewidth}{!}{%
\begin{tabular*}{\linewidth}{@{\extracolsep{\fill}}l *{3}{>{\centering\arraybackslash}p{1.1cm}} c@{}}
        \toprule
        \multirow{2}{*}{\textbf{Setup}} & \multicolumn{3}{c}{\textbf{Token}} & \multirow{2}{*}{\textbf{Avg Score}} \\
        \cmidrule(r{9pt}){2-4}
         & \textbf{Text} & \textbf{Visual} & \textbf{Total} & \\
        \midrule
        NaiveRAG ($K$=10) & 0.8k & -- & 0.8k\costdown{4.3k} & 57.04\gap{19.14} \\
        FC (T) & 9.0k & -- & 9.0k\costup{3.9k} & 34.39\gap{41.79} \\
        \multicolumn{5}{@{}c@{}}{\vspace{-3.5pt}\hdashrule[1.3ex]{\linewidth}{0.35pt}{2.5pt 2.5pt}} \\
        FC (V) & 7.2k & 12.7k & 19.9k\costup{14.8k} & 31.90\gap{44.28} \\
        MuRAG ($K$=5) & 0.3k & 2.0k & 2.3k\costdown{2.8k} & 64.88\gap{11.30} \\
        MuRAG ($K$=10) & 0.6k & 3.9k & 4.5k\costdown{0.6k} & \underline{66.53}\gap{9.65} \\
        MuRAG ($K$=15) & 1.0k & 5.9k & 6.9k\costup{1.8k} & 66.11\gap{10.07} \\
        MuRAG ($K$=20) & 1.3k & 7.8k & 9.1k\costup{4.0k} & 66.09\gap{10.09} \\
        \multicolumn{5}{@{}c@{}}{\vspace{-3.5pt}\hdashrule[1.3ex]{\linewidth}{0.35pt}{2.5pt 2.5pt}} \\
        \rowcolor{RowPurple}
        \textbf{LMMR} & \textbf{4.3k} & \textbf{0.8k} & \textbf{5.1k} & \textbf{76.18} \\
        \bottomrule
    \end{tabular*}}
\end{minipage}
\end{figure}

\paragraph{Better evidence use supports both answering and abstention.}
Fig.~\ref{fig:memgallery_pareto} shows improved capability alongside strong refusal performance, suggesting that gains do not simply reflect answering more often. Memory evidence guides both whether to answer and what to answer.

\paragraph{Efficiency depends on how tokens are used, not only how many are read.}
Table~\ref{tab:efficiency} favors dialogue context and selected images over extensive visual retrieval. Token counts exclude latent computation and cannot establish runtime savings; Appendix~\ref{app:metrics} details the accounting.

\begin{figure}[!t]
\centering
\begin{minipage}[t]{0.48\linewidth}
\vspace{0pt}\centering
\includegraphics[width=\linewidth]{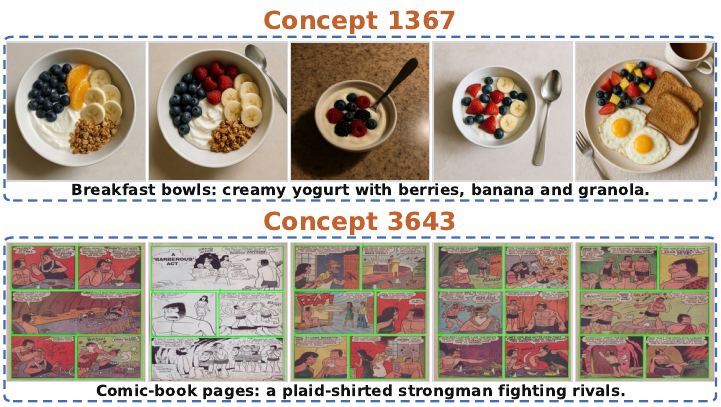}
\caption{Shared SAE concepts on MemEye without dictionary refitting.}
\label{fig:memeye_concept}
\end{minipage}\hfill
\vspace{4mm}
\begin{minipage}[t]{0.49\linewidth}
\vspace{0pt}\centering
\includegraphics[width=\linewidth]{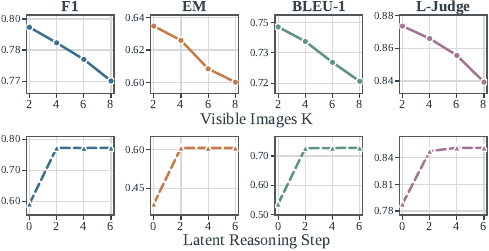}
\caption{Effects of visible-image budget $K$ and the number of latent reasoning steps.}
\label{fig:sweep-sensitivity}
\end{minipage}
\end{figure}

\paragraph{Shared concepts provide reusable addresses for unseen memories.}
The coherent MemEye examples in Fig.~\ref{fig:memeye_concept} suggest that the dictionary retains semantic meaning outside the benchmark on which it was learned. This supports associating memories through common concepts across different descriptions. The dictionary provides semantic addresses for related observations, while native visual tokens preserve the instance-specific details that concept-level abstraction can omit. Quantitative cross-modal alignment analyses for both reader families are provided in Appendix~\ref{app:analyses}.

\paragraph{A short reasoning stage is more useful than a larger visual prefix.}
The contrasting sweeps in Fig.~\ref{fig:sweep-sensitivity} favor allocating computation to evidence integration over increasing visible content. Early saturation suggests that a compact rollout can exploit the selected evidence, whereas additional images may introduce competing details. Within the tested ranges, this supports a selective visual prefix and a short reasoning stage, with further gains depending on better evidence use.

\begin{figure*}[!t] \centering \includegraphics[width=\textwidth]{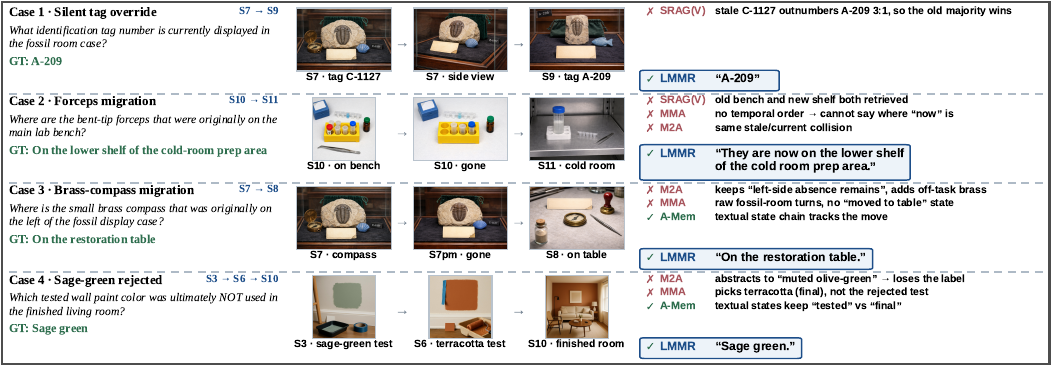} \caption{Case studies on MemEye with Qwen3-VL-8B. Each row shows a question, its ground truth, the decisive visual evidence, what the retrieval baselines answer and what ours answers.} \label{fig:memeye_case_study} \end{figure*}

\paragraph{Temporal validity cannot be inferred from visual similarity alone.}
The cases in Fig.~\ref{fig:memeye_case_study} share a failure mode: an earlier observation remains visually relevant after it has ceased to support the answer. Repeated retrieval can reinforce that stale observation or conflate intermediate and final states. LMMR's responses illustrate the value of interpreting image details in their dialogue chronology. A-Mem's successes in two cases further suggest that resolving state transitions is the essential requirement, with native vision providing additional evidence when textual records are incomplete.

\section{Conclusion}
We presented LMMR, which aligns text and visual features in a shared sparse concept space, organizes evidence through a dialogue-turn graph, and integrates the resulting memory state through a continuous latent rollout. Native visual tokens preserve details that textual summaries may omit. Experiments on Mem-Gallery and zero-shot MemEye evaluation show improved recall, while ablations highlight the complementary roles of dialogue context, visual evidence, and latent reasoning. The sensitivity analysis favors selective visual access and short rollouts over indiscriminately expanding the input. Fine-grained open-ended recall remains challenging; separately isolating the graph and dictionary and measuring end-to-end runtime are useful directions for further evaluation.



\subsection*{AI use statement}
Generative AI tools were used to polish the manuscript and assist in finding relevant research. The authors are responsible for the final text, citation accuracy, and scientific claims. AI models also form part of the method and evaluation pipeline; Appendices~\ref{app:metrics} and~\ref{app:prompts} document the model-based evaluation protocol and prompts used to score generated answers.

\subsection*{Ethics statement}
This work studies multimodal memory using existing benchmarks. Persistent image and dialogue memories can expose sensitive information, retain outdated personal facts, and amplify biases in the underlying models or data. Deployment should therefore provide informed consent, access controls, and mechanisms to inspect, correct, and delete stored memories, while respecting dataset licenses and usage conditions. Benchmark improvements do not establish reliability in high-stakes settings; model-judged scores may also reflect evaluator bias.

\subsection*{Reproducibility statement}
The \hyperref[app:contents]{appendix contents} (p.~\pageref{app:contents}) link to all supplementary sections. Section~\ref{sec:method} describes the architecture and training objective. Appendices~\ref{app:benchmarks}--\ref{app:setup} document the benchmarks, evaluation metrics, scenario-level splits, leakage controls, and key inference hyperparameters; Algorithm~\ref{alg:inference} specifies the inference procedure. Appendices~\ref{app:memgallery-full} and~\ref{app:memeye-full} report complete result matrices, and Appendix~\ref{app:prompts} provides the answer-generation and judge prompts used in the reported evaluations.

\setcitestyle{numbers,square}
\bibliography{iclr2027_conference}

\clearpage
\appendix
\section*{Appendix}
\label{app:contents}
\pdfbookmark[0]{Appendix Contents}{appendixcontents}
\begingroup
\setlength{\parskip}{5pt}
\newcommand{\appcontentsline}[3]{\noindent\hspace*{#1em}\hyperref[#2]{\ref*{#2}\quad #3}\nobreak\dotfill\pageref{#2}\par}
\smallskip
\appcontentsline{0}{sec:appendix}{Experimental Details and Reproducibility}
\appcontentsline{1}{app:benchmarks}{Datasets and benchmarks}
\appcontentsline{2}{app:memgallery}{Mem-Gallery}
\appcontentsline{2}{app:memeye}{MemEye}
\appcontentsline{1}{app:metrics}{Evaluation protocol}
\appcontentsline{1}{app:setup}{Implementation details}
\smallskip
\appcontentsline{0}{app:results}{Complete Result Matrices}
\appcontentsline{1}{app:memgallery-full}{Per-task results on Mem-Gallery}
\appcontentsline{1}{app:memeye-full}{Per-setting results on MemEye}
\smallskip
\appcontentsline{0}{app:analyses}{Additional analyses}
\appcontentsline{1}{app:alignment}{Cross-modal alignment of the shared concept space}
\appcontentsline{1}{app:probe}{A quantitative alignment probe}
\smallskip
\appcontentsline{0}{app:prompts}{Prompts}
\appcontentsline{1}{app:prompt-answer}{Answer generation}
\appcontentsline{1}{app:prompt-judge}{Judge prompts}
\endgroup
\clearpage
\section{Experimental Details and Reproducibility}
\label{sec:appendix}

\providecolor{AppHeader}{RGB}{242, 242, 242}
\providecolor{AppTint}{RGB}{238, 234, 248}

\subsection{Datasets and benchmarks}
\label{app:benchmarks}

The two benchmarks we use are complementary in what they hold constant. Mem-Gallery keeps the task broad and the history long, and asks whether a memory system can be organised well enough to serve nine different question types; MemEye keeps the task small and varies the evidence itself, from scene to pixel, so that a system's failure can be attributed to a granularity level rather than to a topic. We describe each release as we consume it, and recompute the dataset statistics from the public releases using the same history boundaries and evaluation protocol as our experiments.

\subsubsection{Mem-Gallery}
\label{app:memgallery}

\noindent\textbf{What it contains.} Twenty persona-driven conversation scenarios, each a multi-session dialogue in which images are inserted at information-relevant rounds, plus human-annotated question--answer (QA) pairs that record the clue rounds supporting each answer. Table~\ref{tab:app-mg-overview} summarises the release along the axes our memory pipeline actually reads: scenarios, sessions, rounds, images and questions. Two numbers are worth separating here. The image repository ships 1{,}490 files, but only 1{,}003 of them are attached to a dialogue round; the remaining 487 files are never referenced by any turn and are unused in our pipeline. Likewise, a question does not see the whole history: our preprocessing resolves every clue round to a round index and defines the \emph{anchor} of a question as the last session it references, so that all later rounds are treated as future information and withheld. This makes retrieval-based and state-based memories comparable under a single, well-posed recall setting.

\noindent\textbf{Annotation structure.} Each QA pair carries a task label, a short answer, the sessions it depends on and up to sixteen clue rounds. Two thirds of the questions (1{,}130, 66.0\%) require looking at an image, and 117 of them (6.8\%) can only be answered by combining evidence from more than one session. The decisive evidence is usually not adjacent to the question either: a question sits 115.8 rounds after the start of its scenario on average, and the memory it has to organise spans 73.5 days on average (median 68, maximum 216). This is the regime LMMR is designed for---evidence that is distant, distributed across sessions, and often dependent on visual details.

The nine categories fall into three families that stress different stages of a memory system:

\begin{table}[!t]
\centering
\caption{\textbf{Mem-Gallery at a glance.} Statistics recomputed from the public release. ``Visual-evidence QA'' counts questions whose supporting clue rounds carry at least one image; clue statistics are reported over the 1{,}527 answerable questions (the 184 refusal items have no supporting clue by construction). History and span are measured up to the anchor of each question.}
\label{tab:app-mg-overview}
\vspace{2pt}
\small
\setlength{\tabcolsep}{0.6em}
\renewcommand{\arraystretch}{1.10}
\begin{tabular}{@{}l r@{}}
\toprule
\rowcolor{AppHeader}
\textbf{Property} & \textbf{Value} \\
\midrule
Conversation scenarios & 20 \\
Sessions (per scenario) & 240 (12.0) \\
Dialogue rounds (per session) & 3{,}962 (16.5) \\
Images attached to rounds (per session) & 1{,}003 (4.2) \\
Image files in the release (unreferenced) & 1{,}490 (487) \\
Annotated QA pairs (per scenario) & 1{,}711 (85.6) \\
Average answer length & 6.0 words \\
QA with visual evidence & 1{,}130 (66.0\%) \\
QA spanning more than one session & 117 (6.8\%) \\
Clue rounds per answerable QA (median / max) & 2.69 (2 / 16) \\
Clue rounds spanning more than one session & 115 (7.5\%) \\
History preceding a question (median) & 115.8 rounds (114) \\
Temporal span covered by the memory (median) & 73.5 days (68) \\
\bottomrule
\end{tabular}
\end{table}

\begin{itemize}
\setlength{\itemsep}{1pt}
\item \textit{Memory extraction and test-time adaptation.} \textbf{FR} (factual retrieval) recovers a single stated fact; \textbf{VS} (visual-centric search) has to locate \emph{which} image of the history the question refers to, and its gold answer is the image identifier itself; \textbf{TTL} (test-time learning) applies a convention or skill that is introduced only inside the history.
\item \textit{Memory reasoning.} \textbf{TR} (temporal reasoning) orders, dates or compares events across sessions; \textbf{VR} (visual-centric reasoning) infers a conclusion the dialogue text never restates; \textbf{MR} (multi-entity reasoning) combines evidence about several entities at once.
\item \textit{Knowledge management.} \textbf{KR} (knowledge resolution) keeps the currently valid fact after an update; \textbf{CD} (conflict detection) notices that two observations contradict; \textbf{AR} (answer refusal) asks for abstention when the evidence does not exist---by construction these items carry no clue, so their averages are reported as ``---''.
\end{itemize}

Table~\ref{tab:app-mg-tasks} shows how the annotation mass is distributed. TTL and VS dominate (19.7\% and 17.9\% of all questions), while KR and CD are the smallest categories at 4.7\% each. The organising load is very uneven: TTL questions need 4.09 clue rounds on average and 85.2\% of them need at least three, whereas CD averages 1.70 and TR 1.74. TR is the most temporal category in the cross-session sense---35.8\% of its questions depend on evidence from more than one session, against 0.5\% for FR---and CD and KR ask for the shortest answers.

\begin{table}[t]
\centering
\caption{\textbf{Mem-Gallery task statistics.} Share is relative to all 1{,}711 questions. ``$\geq$3 clues'' is the fraction of questions whose answer needs at least three clue rounds; ``Cross-session'' is the fraction whose clues span more than one session; ``Visual'' is the fraction supported by one or more images.}
\label{tab:app-mg-tasks}
\vspace{2pt}
\small
\setlength{\tabcolsep}{0.55em}
\renewcommand{\arraystretch}{1.10}
\begin{tabular}{@{}l r r r r r r@{}}
\toprule
\rowcolor{AppHeader}
\textbf{Task} & \textbf{\#QA} & \textbf{Share (\%)} & \textbf{Avg. clues} & \textbf{$\geq$3 clues (\%)} & \textbf{Cross-session (\%)} & \textbf{Visual (\%)} \\
\midrule
FR  & 219 & 12.8 & 2.90 & 35.2 & 0.5  & 51.6 \\
VS  & 306 & 17.9 & 1.91 & 24.2 & 6.9  & 100.0 \\
TTL & 337 & 19.7 & 4.09 & 85.2 & 9.5  & 100.0 \\
TR  & 123 & 7.2  & 1.74 & 10.6 & 35.8 & 35.0 \\
VR  & 174 & 10.2 & 3.05 & 54.0 & 4.0  & 99.4 \\
MR  & 206 & 12.0 & 2.17 & 19.9 & 1.9  & 43.2 \\
KR  & 81  & 4.7  & 2.21 & 34.6 & 0.0  & 45.7 \\
CD  & 81  & 4.7  & 1.70 & 16.0 & 7.4  & 39.5 \\
AR  & 184 & 10.8 & ---  & ---  & ---  & 0.0 \\
\midrule
\rowcolor{AppTint}
\textbf{Overall} & \textbf{1{,}711} & \textbf{100} & \textbf{2.69} & \textbf{36.6} & \textbf{6.8} & \textbf{66.0} \\
\bottomrule
\end{tabular}
\end{table}

\noindent\textbf{Scenario balance.} Table~\ref{tab:app-mg-scenarios} lists the twenty scenarios individually. They differ substantially in size, from 9 to 18 sessions (144 to 270 rounds), 30 to 117 dialogue images and 54 to 205 questions, and the visual density varies as well: in the Programming scenario 76.1\% of the questions have visual evidence, in the Astronomy scenario only 46.3\%. A question-level split would therefore leak scenario-specific writing style into training; we cross-fit at the scenario level instead, splitting the twenty scenarios into five folds of four (fold sizes 278, 294, 339, 369 and 431 held-out questions), so that the evaluation scenarios of a fold are never seen during training of that fold.

\begin{table}[t]
\centering
\caption{\textbf{Mem-Gallery per-scenario statistics.} Scenario names are abbreviated. ``Img.'' counts the images attached to dialogue rounds (1{,}003 in total across the release); ``VisQA'' counts questions supported by at least one image in their annotated clue rounds.}
\label{tab:app-mg-scenarios}
\vspace{2pt}
\footnotesize
\setlength{\tabcolsep}{0.18em}
\renewcommand{\arraystretch}{1.05}
\begin{tabular}{@{}l r r r r r @{\hspace{0.9em}} l r r r r r@{}}
\toprule
\rowcolor{AppHeader}
\textbf{Scenario} & \textbf{Sess.} & \textbf{Rounds} & \textbf{Img.} & \textbf{QA} & \textbf{VisQA}
& \textbf{Scenario} & \textbf{Sess.} & \textbf{Rounds} & \textbf{Img.} & \textbf{QA} & \textbf{VisQA} \\
\midrule
AI, Robotics        & 12 & 185 & 31  & 57  & 35  & Travel, Fashion    & 9  & 159 & 53  & 78  & 59 \\
Academic, Pets      & 14 & 177 & 45  & 77  & 53  & Home, Health       & 10 & 163 & 36  & 74  & 41 \\
Architecture, Art   & 14 & 186 & 30  & 74  & 38  & Home Repair        & 18 & 270 & 43  & 68  & 35 \\
Astronomy, Physics  & 11 & 203 & 30  & 54  & 25  & Landscape, Nature  & 11 & 173 & 64  & 92  & 73 \\
Baking, Desserts    & 15 & 262 & 57  & 111 & 71  & Music, Theater     & 14 & 220 & 43  & 57  & 30 \\
Dog Behavior        & 11 & 180 & 38  & 63  & 38  & Programming        & 10 & 201 & 78  & 136 & 103 \\
Education, Career   & 12 & 227 & 56  & 106 & 65  & Parenting, Gear    & 16 & 221 & 48  & 82  & 54 \\
Entrepreneurship    & 10 & 144 & 36  & 90  & 53  & Python, Botany     & 12 & 252 & 117 & 205 & 156 \\
Fashion, Shopping   & 9  & 168 & 37  & 61  & 40  & Real Estate        & 11 & 196 & 53  & 75  & 48 \\
Global Travel       & 10 & 176 & 66  & 87  & 62  & Tech, Ethics       & 11 & 199 & 42  & 64  & 51 \\
\midrule
\rowcolor{AppTint}
\textbf{Subtotal} & \textbf{118} & \textbf{1{,}908} & \textbf{426} & \textbf{780} & \textbf{480}
& \textbf{Subtotal} & \textbf{122} & \textbf{2{,}054} & \textbf{577} & \textbf{931} & \textbf{650} \\
\bottomrule
\end{tabular}
\end{table}

\subsubsection{MemEye}
\label{app:memeye}

\noindent\textbf{What it contains.} MemEye~\citep{guo2026memeye} is organised the other way round: instead of tasks, its questions are laid out on a coordinate system whose axes are the granularity of the decisive visual evidence ($X1$ scene, $X2$ region, $X3$ instance, $X4$ pixel) and the depth of reasoning over memory ($Y1$ atomic retrieval, $Y2$ relational association, $Y3$ evolutionary synthesis). Eight life-scenario tasks in four domains contribute 221 sessions, 848 dialogue rounds and 438 images; every question is written in two mirrored forms, so the 371 annotated items correspond to 371 open-ended questions and 1{,}484 multiple-choice items after the four-way option rotation. Table~\ref{tab:app-me-tasks} gives the per-task breakdown, and two properties of the release shape how we use it: images are sparse relative to dialogue (438 of the 848 rounds carry one, and the density per task ranges from 29.1\% to 100\%), and questions are long enough to be self-contained---22.1 words on average, up to 79---while gold answers are short (5.6 words on average in the open-ended form), so the benchmark rewards precise recall rather than fluent description. On average a question needs 2.81 clue rounds (median 2, maximum 43) and 46.1\% of the questions draw on more than one session; those multi-session items sit mostly at the relational level (137 of the 171 are $Y2$), so most cross-session questions require combining related observations rather than retrieving a single fact from an isolated dialogue turn.

\begin{table}[t]
\centering
\caption{\textbf{MemEye per-task statistics.} Domains follow the original release (Leisure, Domestic, Professional, Personal). ``Img.'' counts the distinct images referenced by dialogue rounds (438 of the 495 image files shipped with the release); each of these rounds carries exactly one image, so the last column is the share of rounds that are image-grounded.}
\label{tab:app-me-tasks}
\vspace{2pt}
\small
\setlength{\tabcolsep}{0.55em}
\renewcommand{\arraystretch}{1.10}
\begin{tabular}{@{}l l r r r r r@{}}
\toprule
\rowcolor{AppHeader}
\textbf{Task} & \textbf{Domain} & \textbf{Sessions} & \textbf{Rounds} & \textbf{QA} & \textbf{Img.} & \textbf{Img. rounds (\%)} \\
\midrule
Brand Memory Test                & Professional & 42 & 72  & 29 & 30 & 41.7 \\
Card Playlog Test                & Leisure      & 4  & 30  & 48 & 30 & 100.0 \\
Cartoon Entertainment Companion  & Leisure      & 86 & 299 & 76 & 87 & 29.1 \\
Home Renovation, Interior Design & Domestic     & 13 & 120 & 52 & 89 & 74.2 \\
Multi-Scene Visual Case Archive  & Professional & 15 & 117 & 50 & 57 & 48.7 \\
Outdoor Navigation Route Memory  & Domestic     & 10 & 60  & 28 & 40 & 66.7 \\
Personal Health Dashboard        & Personal     & 12 & 97  & 51 & 62 & 63.9 \\
Social Chat Memory               & Personal     & 39 & 53  & 37 & 43 & 81.1 \\
\midrule
\rowcolor{AppTint}
\textbf{Total} & \textbf{4 domains} & \textbf{221} & \textbf{848} & \textbf{371} & \textbf{438} & \textbf{51.7} \\
\bottomrule
\end{tabular}
\end{table}

\noindent\textbf{The $(X,Y)$ grid.} Table~\ref{tab:app-me-grid} is the annotation grid of the benchmark and, read literally, the source of the twelve settings of the main text: each cell is one $(X,Y)$ combination, and the per-setting curves of Fig.~\ref{fig:memeye} aggregate questions by cell. The mass is concentrated towards fine-grained evidence and shallow-to-mid reasoning---$X3$ and $X4$ hold 275 of the 371 questions (74.1\%), the $Y2$ column alone holds 204 (55.0\%), and the largest single cell is $X4/Y2$ with 87 questions, i.e.\ pixel-level evidence that has to be associated across observations. The deepest cells are also the thinnest: $X1/Y3$ has 8 questions and $X2/Y3$ has 5, so cell-level differences there are indicative rather than conclusive, and we therefore report per-axis aggregates alongside the per-cell ones. The imbalance is deliberate---it is what makes the benchmark discriminate between captions and pixels.

\begin{table}[t]
\centering
\caption{\textbf{MemEye annotation grid.} Number of questions per $(X,Y)$ cell, with margins and shares. Rows represent the granularity of the decisive visual evidence ($X1$ scene, $X2$ region, $X3$ instance, $X4$ pixel), columns represent the depth of reasoning over memory ($Y1$ atomic retrieval, $Y2$ relational association, $Y3$ evolutionary synthesis). The twelve cells are the twelve settings of the main text.}
\label{tab:app-me-grid}
\vspace{2pt}
\small
\setlength{\tabcolsep}{0.9em}
\renewcommand{\arraystretch}{1.12}
\begin{tabular}{@{}l r r r r r@{}}
\toprule
\rowcolor{AppHeader}
\textbf{Evidence $\backslash$ Reasoning} & \textbf{$Y1$} & \textbf{$Y2$} & \textbf{$Y3$} & \textbf{Total} & \textbf{Share (\%)} \\
\midrule
$X1$ scene     & 10 & 30 & 8  & 48  & 12.9 \\
$X2$ region    & 15 & 28 & 5  & 48  & 12.9 \\
$X3$ instance  & 74 & 59 & 9  & 142 & 38.3 \\
$X4$ pixel     & 20 & 87 & 26 & 133 & 35.8 \\
\midrule
\rowcolor{AppTint}
\textbf{Total} & \textbf{119} & \textbf{204} & \textbf{48} & \textbf{371} & \textbf{100} \\
\textbf{Share (\%)} & 32.1 & 55.0 & 12.9 & 100 & --- \\
\bottomrule
\end{tabular}
\end{table}

\subsection{Evaluation protocol}
\label{app:metrics}

\noindent\textbf{Metrics.} On Mem-Gallery we report token-level F1, BLEU-1, exact match (EM) and an LLM-as-a-Judge (L-Judge) score obtained with Qwen2.5-72B-Instruct under the official prompt, and we keep all four rather than averaging them away: F1 and BLEU-1 reward partial overlap, EM requires the answer to be exactly right (including image identifiers in VS), and the judge is the only metric that tolerates paraphrases. On MemEye, multiple-choice questions are scored by exact match averaged over the four answer rotations, which cancels position bias, and open-ended questions are scored by GPT-5.2 under the official judge prompt~\citep{guo2026memeye}.

\noindent\textbf{Splits and leakage control.} Every Mem-Gallery question is answered against its own anchor only, so no system can see the sessions that a question does not reference; the five scenario-level folds above then give each question a prediction from a model that never trained on its scenario, and the numbers we report aggregate the held-out items of all folds, with the fold-to-fold spread reported as a standard deviation. MemEye is evaluated zero-shot with the Mem-Gallery checkpoint: no MemEye question, session or image is used during training, which is why we can read its per-granularity curves as a transfer test rather than as an in-domain result. For settings with fewer than ten questions, we describe broad trends without treating score gaps as reliable differences.

\noindent\textbf{Derived artefacts.} For the token accounting of Table~\ref{tab:efficiency} we tokenise dialogue rounds at 64 tokens and image captions at 56 tokens per entry, and measure a retrieved image at 391 visual tokens; the memory context is capped at 6{,}144 tokens throughout. The shared dictionary is fitted once on the Mem-Gallery memory bank and then frozen, so MemEye evaluates transfer through the existing concept space without adapting the dictionary to the target benchmark.

\subsection{Implementation details}
\label{app:setup}

Unless stated otherwise, the shared dictionary holds $C{=}4096$ entries with top-$k{=}32$ activations, the dialogue-turn graph uses $L{=}2$ message-passing layers, the latent rollout runs for $S{=}10$ steps with a single graph injection, $K{=}2$ images enter the multimodal prefix, and the memory context is capped at 6{,}144 tokens.
Hyperparameters are selected once on a single fold and then kept fixed for the remaining folds.
Algorithm~\ref{alg:inference} states the inference path under these settings. Here, LM denotes the language model, MHSA denotes multi-head self-attention, and MLP denotes a multilayer perceptron; the remaining symbols follow the method notation defined in Section~\ref{sec:method}.

\begin{algorithm}[!t]
\centering
\small
\setlength{\tabcolsep}{3pt}
\renewcommand{\arraystretch}{1.15}
\begin{tabular}{@{}r@{\hspace{0.7em}}p{0.88\textwidth}@{}}
\toprule
\multicolumn{2}{@{}l}{\textit{Indexing (once per memory bank)}} \\
\midrule
1: & \textbf{for} each dialogue turn $u$, its caption $c_u$ and its image $I_u$ \textbf{do} \\
2: & \quad $z_u, z_u^{c} \leftarrow \mathrm{TopK}\big(\langle W_t h_u, D\rangle\big)$ \\
3: & \quad $z_u^{v} \leftarrow \mathrm{TopK}\big(\langle W_v \mathbf{U}_u, D\rangle\big)$ \hfill {\footnotesize ($k=32$ entries of $C=4096$)} \\
4: & \quad store $z_u, z_u^{c}, z_u^{v}$, the native visual tokens $\mathbf{R}_u$ and the provenance $(\tau_u, \mathrm{id}_u)$ \\
\midrule
\multicolumn{2}{@{}l}{\textit{Answering a query $q$}} \\
\midrule
5: & $z_q \leftarrow \mathrm{TopK}\big(\langle W_t h_q, D\rangle\big)$ \\
6: & $G_q \leftarrow$ nodes whose code shares an entry with $z_q$, plus their temporal neighbours \\
7: & keep the $\le K$ images with the strongest shared activations, ranked by activation \\
8: & \textbf{for} $\ell = 1 \dots L$ \textbf{do} \hfill {\footnotesize} \\
9: & \quad $H \leftarrow H + \mathrm{MHSA}(H, A_q)$; \quad $H \leftarrow H + \mathrm{MLP}(H)$ \\
10: & $r \leftarrow \mathrm{Readout}(H)$ \hfill {\footnotesize (graph summary in concept space)} \\
11: & $h \leftarrow \mathrm{LM}_{\mathrm{prefix}}\big(G_q, \{\mathbf{R}_u\}_{u \in G_q}\big)$; \quad $h \leftarrow h + W_o r$ \hfill {\footnotesize (single residual injection)} \\
12: & \textbf{for} $s = 1 \dots S$ \textbf{do} \quad $h \leftarrow \mathrm{LM}_{\mathrm{step}}(h, r)$ \hfill {\footnotesize} \\
13: & \textbf{return} $a \leftarrow \mathrm{Decode}(h)$ \\
\bottomrule
\end{tabular}
\caption{\textbf{Inference with LMMR.} Steps 1--4 run once when the bank is built, steps 5--13 once per query; no parameter is updated at inference time.}
\label{alg:inference}
\end{algorithm}

\section{Complete Result Matrices}
\label{app:results}

\subsection{Per-task results on Mem-Gallery}
\label{app:memgallery-full}
\providecolor{HeaderLavender}{HTML}{E9E3F2}
\providecolor{BestOrange}{HTML}{FCE4D6}
\providecolor{SecondBlue}{HTML}{DDEBF7}
\providecommand{\best}[1]{\cellcolor{BestOrange}\textbf{#1}}
\providecommand{\runnerup}[1]{\cellcolor{SecondBlue}\textbf{#1}}

Table~\ref{tab:memgallery_full_comparison} opens up the comparison of Table~\ref{tab:memgallery_overall_size_comparison}: the nine task categories are rows, the four metrics are sub-rows, and the columns are the thirteen published systems together with LMMR, all read from the same five folds. Read column-wise, LMMR is the strongest system on eight of the nine categories in F1---on refusal it ties FIFO---and second only on visual-centric search, where MuRAG keeps a 1.0-point edge with an image retriever that reads the whole history rather than a memory state. The size of the margin is, however, far from uniform. Where a question can be answered by matching one retrieved passage, the two families of systems sit within a few points: visual-centric search (87.15 vs.\ 88.18), multi-entity reasoning (50.39 vs.\ 50.07) and refusal (100.00 vs.\ 100.00). The categories that require several observations to be ordered, updated or complemented are the ones where the published systems plateau: conflict detection moves from a band of 32.10--37.04 shared by every baseline to 86.42, temporal reasoning from 58.33 to 85.71, and test-time learning improves from 81.77 to 89.79 under the same evaluation protocol.

Two properties of the table are worth naming instead of glossing over. Conflict detection is scored as a binary decision, so its four sub-rows coincide for every system, ours included; the informative quantity there is the gap to the plateau, not the choice of metric. Refusal behaves in the opposite way: abstention is easy to score well on, and FIFO, FC (T) and Reflexion reach 99.58--100.00 F1 precisely because they refuse almost always, which is why the main text reads that column together with capability rather than in isolation. Our own refusal row comes from a system that also answers the 1,527 answerable questions at the best F1 in the table and abstains on none of them.

\begin{table*}[!t]
\centering
\scriptsize
\setlength{\tabcolsep}{2.5pt}
\renewcommand{\arraystretch}{1.02}
\caption{Per-task comparison on Mem-Gallery with the Qwen-2.5-VL-7B backbone.}
\resizebox{\textwidth}{!}{%
\begin{tabular}{@{}ccc*{8}{r}!{\vrule width 0.5pt}*{6}{r}@{}}
\toprule
\rowcolor{HeaderLavender}
\multicolumn{3}{c}{\textbf{Qwen-2.5-VL-7B}}
& \shortstack{FC (T)}
& FIFO
& \shortstack{Naive RAG}
& \shortstack{Gen.Agent}
& Reflexion
& MemGPT
& A-Mem
& MemoryOS
& \shortstack{FC (V)}
& MuRAG
& \shortstack{Universal RAG}
& NGM
& AUGUSTUS
& \textbf{LMMR (Ours)} \\
\midrule
\rowcolor{white}

\multirow{12}{*}{\rotatebox[origin=c]{90}{\scriptsize Extract.\ \& Adapt.}}
& \multirow{4}{*}{FR}
& F1
& 0.2376 & 0.1446 & 0.5852 & 0.2424 & 0.2391 & 0.5928 & 0.6072 & 0.6244
& 0.2150 & \runnerup{0.6724} & 0.6632 & 0.6364 & 0.6162 & \best{0.7366} \\
&
& BLEU-1
& 0.1865 & 0.1000 & 0.5045 & 0.1862 & 0.1903 & 0.5098 & 0.5138 & 0.5354
& 0.1626 & \runnerup{0.5755} & 0.5658 & 0.5600 & 0.5309 & \best{0.6453} \\
&
& EM
& 0.0913 & 0.0457 & 0.3059 & 0.0868 & 0.0913 & 0.3288 & 0.2922 & 0.3470
& 0.0685 & 0.3607 & 0.3470 & \runnerup{0.3744} & 0.3288 & \best{0.4110} \\
&
& LLM-Judge
& 0.2626 & 0.1324 & 0.7763 & 0.2945 & 0.2626 & 0.8539 & 0.7808 & 0.8265
& 0.2260 & \runnerup{0.8790} & 0.8744 & 0.8082 & 0.8082 & \best{0.8973} \\
\cmidrule(lr){2-17}

& \multirow{4}{*}{VS}
& F1
& 0.1992 & 0.0612 & 0.7592 & 0.2970 & 0.1954 & 0.6239 & 0.7681 & 0.7853
& 0.1658 & \best{0.8818} & 0.8708 & 0.8531 & 0.8499 & \runnerup{0.8715} \\
&
& BLEU-1
& 0.1873 & 0.0549 & 0.7063 & 0.2616 & 0.1840 & 0.5834 & 0.6880 & 0.7181
& 0.1473 & \best{0.8442} & 0.8343 & 0.8145 & 0.7942 & \runnerup{0.8381} \\
&
& EM
& 0.1601 & 0.0392 & 0.5686 & 0.2124 & 0.1569 & 0.4118 & 0.5882 & 0.6046
& 0.1078 & 0.6699 & 0.6699 & \runnerup{0.6863} & 0.6536 & \best{0.7157} \\
&
& LLM-Judge
& 0.1961 & 0.0556 & 0.7402 & 0.2958 & 0.1895 & 0.5964 & 0.7369 & 0.7729
& 0.1683 & \best{0.8856} & \runnerup{0.8611} & 0.8480 & 0.8399 & 0.8513 \\
\cmidrule(lr){2-17}

& \multirow{4}{*}{TTL}
& F1
& 0.4500 & 0.3351 & 0.6526 & 0.4851 & 0.4486 & 0.2924 & 0.6336 & 0.5484
& 0.4147 & \runnerup{0.8177} & 0.7824 & 0.7817 & 0.7913 & \best{0.8979} \\
&
& BLEU-1
& 0.3799 & 0.2692 & 0.5835 & 0.4150 & 0.3798 & 0.2295 & 0.5595 & 0.4697
& 0.3477 & \runnerup{0.7449} & 0.7103 & 0.7168 & 0.7206 & \best{0.8843} \\
&
& EM
& 0.2374 & 0.1365 & 0.4718 & 0.2967 & 0.2374 & 0.1009 & 0.4362 & 0.3561
& 0.2107 & \runnerup{0.6172} & 0.6142 & 0.6024 & 0.6113 & \best{0.8427} \\
&
& LLM-Judge
& 0.7092 & 0.6677 & 0.8457 & 0.7582 & 0.7033 & 0.7092 & 0.7997 & 0.7715
& 0.7107 & 0.9006 & 0.8501 & \runnerup{0.9110} & 0.8932 & \best{0.9228} \\
\cmidrule(lr){2-17}

\multirow{12}{*}{\rotatebox[origin=c]{90}{\scriptsize Reasoning}}
& \multirow{4}{*}{TR}
& F1
& 0.2545 & 0.1549 & 0.4887 & 0.2742 & 0.2553 & 0.5661 & 0.5604 & 0.5497
& 0.2294 & \runnerup{0.5833} & 0.5460 & 0.5425 & 0.5800 & \best{0.8571} \\
&
& BLEU-1
& 0.2363 & 0.1316 & 0.4587 & 0.2473 & 0.2363 & 0.5326 & 0.5361 & 0.5240
& 0.2065 & \runnerup{0.5537} & 0.5137 & 0.5143 & 0.5527 & \best{0.8242} \\
&
& EM
& 0.1545 & 0.0894 & 0.3496 & 0.1626 & 0.1545 & 0.3496 & 0.4065 & 0.3821
& 0.1382 & 0.4309 & 0.4065 & 0.4146 & \runnerup{0.4390} & \best{0.7236} \\
&
& LLM-Judge
& 0.2805 & 0.1463 & 0.6382 & 0.3252 & 0.2764 & \runnerup{0.8008} & 0.6951 & 0.7195
& 0.2480 & 0.7724 & 0.7520 & 0.7195 & 0.7398 & \best{0.9187} \\
\cmidrule(lr){2-17}

& \multirow{4}{*}{VR}
& F1
& 0.2552 & 0.1207 & 0.3022 & 0.1955 & 0.2594 & 0.4593 & 0.4477 & 0.4280
& 0.2015 & 0.4818 & \runnerup{0.4879} & 0.4615 & 0.3866 & \best{0.6225} \\
&
& BLEU-1
& 0.2442 & 0.1005 & 0.2873 & 0.1815 & 0.2480 & 0.4459 & 0.4331 & 0.4126
& 0.1912 & 0.4625 & \runnerup{0.4682} & 0.4428 & 0.3726 & \best{0.6002} \\
&
& EM
& 0.2011 & 0.0690 & 0.1954 & 0.1207 & 0.2011 & \runnerup{0.3851} & 0.3563 & 0.3391
& 0.1609 & 0.3793 & 0.3793 & 0.3678 & 0.2931 & \best{0.5402} \\
&
& LLM-Judge
& 0.3046 & 0.1408 & 0.3793 & 0.2471 & 0.3046 & \runnerup{0.6149} & 0.5747 & 0.5805
& 0.2586 & 0.6092 & 0.5977 & 0.5460 & 0.4856 & \best{0.7155} \\
\cmidrule(lr){2-17}

& \multirow{4}{*}{MR}
& F1
& 0.2411 & 0.1745 & 0.4640 & 0.2450 & 0.2428 & 0.4367 & 0.5000 & 0.4490
& 0.2101 & \runnerup{0.5007} & 0.4746 & 0.4866 & 0.4866 & \best{0.5039} \\
&
& BLEU-1
& 0.1739 & 0.1184 & 0.3543 & 0.1784 & 0.1770 & 0.3347 & \best{0.3908} & 0.3446
& 0.1429 & \runnerup{0.3903} & 0.3868 & 0.3655 & 0.3778 & 0.3723 \\
&
& EM
& 0.0340 & 0.0146 & 0.0874 & 0.0243 & 0.0340 & 0.0631 & 0.0728 & 0.0728
& 0.0194 & 0.0874 & \runnerup{0.0971} & 0.0728 & 0.0825 & \best{0.1019} \\
&
& LLM-Judge
& 0.2985 & 0.1602 & 0.7791 & 0.3350 & 0.3058 & 0.8204 & 0.8083 & 0.8204
& 0.2791 & \best{0.8447} & \runnerup{0.8422} & 0.7840 & 0.8228 & 0.8374 \\
\cmidrule(lr){2-17}

\multirow{12}{*}{\rotatebox[origin=c]{90}{\scriptsize Knowledge Management}}
& \multirow{4}{*}{CD}
& F1
& 0.3457 & 0.3580 & 0.3457 & 0.3210 & 0.3333 & 0.3580 & 0.3333 & 0.3333
& 0.3580 & \runnerup{0.3704} & 0.3457 & 0.3457 & 0.3210 & \best{0.8642} \\
&
& BLEU-1
& 0.3457 & 0.3580 & 0.3457 & 0.3210 & 0.3333 & 0.3580 & 0.3333 & 0.3333
& 0.3580 & \runnerup{0.3704} & 0.3457 & 0.3457 & 0.3210 & \best{0.8642} \\
&
& EM
& 0.3457 & 0.3580 & 0.3457 & 0.3210 & 0.3333 & 0.3580 & 0.3333 & 0.3333
& 0.3580 & \runnerup{0.3704} & 0.3457 & 0.3457 & 0.3210 & \best{0.8642} \\
&
& LLM-Judge
& 0.3457 & 0.3580 & 0.3457 & 0.3210 & 0.3333 & 0.3580 & 0.3333 & 0.3333
& 0.3580 & \runnerup{0.3704} & 0.3457 & 0.3457 & 0.3210 & \best{0.8642} \\
\cmidrule(lr){2-17}

& \multirow{4}{*}{KR}
& F1
& 0.2354 & 0.1697 & 0.3595 & 0.2674 & 0.2354 & 0.4292 & 0.4515 & \runnerup{0.5031}
& 0.2181 & 0.4601 & 0.4336 & 0.3944 & 0.3752 & \best{0.6331} \\
&
& BLEU-1
& 0.2005 & 0.1424 & 0.2986 & 0.2341 & 0.2000 & 0.3741 & 0.3992 & \runnerup{0.4479}
& 0.1883 & 0.4073 & 0.3772 & 0.3444 & 0.3159 & \best{0.5845} \\
&
& EM
& 0.1235 & 0.0864 & 0.1605 & 0.1235 & 0.1235 & 0.2099 & 0.2469 & \runnerup{0.2716}
& 0.1235 & 0.2593 & 0.2099 & 0.2099 & 0.1728 & \best{0.4321} \\
&
& LLM-Judge
& 0.3395 & 0.2469 & 0.6358 & 0.3457 & 0.3395 & 0.6667 & 0.7593 & \runnerup{0.7840}
& 0.2840 & 0.7222 & 0.6728 & 0.6728 & 0.6728 & \best{0.8272} \\
\cmidrule(lr){2-17}

& \multirow{4}{*}{AR}
& F1
& \runnerup{0.9958} & \best{1.0000} & 0.9581 & 0.9841 & \runnerup{0.9958} & 0.9849 & 0.9278 & 0.9845
& 0.9946 & 0.9418 & 0.9473 & 0.9580 & 0.9460 & \best{1.0000} \\
&
& BLEU-1
& \runnerup{0.9953} & \best{1.0000} & 0.9575 & 0.9839 & \runnerup{0.9953} & 0.9844 & 0.9257 & 0.9841
& 0.9946 & 0.9409 & 0.9466 & 0.9574 & 0.9459 & \best{1.0000} \\
&
& EM
& \runnerup{0.9946} & \best{1.0000} & 0.9565 & 0.9837 & \runnerup{0.9946} & 0.9837 & 0.9239 & 0.9837
& \runnerup{0.9946} & 0.9402 & 0.9457 & 0.9565 & 0.9457 & \best{1.0000} \\
&
& LLM-Judge
& \runnerup{0.9783} & \best{0.9837} & 0.9402 & 0.9674 & \runnerup{0.9783} & 0.9674 & 0.9375 & 0.9674
& \best{0.9837} & 0.9375 & 0.9429 & 0.9457 & 0.9429 & 0.9755 \\
\cmidrule(lr){2-17}

\multirow{4}{*}{}
& \multirow{4}{*}{Overall}
& F1
& 0.3625 & 0.2724 & 0.5974 & 0.3825 & 0.3619 & 0.5282 & 0.6228 & 0.6109
& 0.3354 & \runnerup{0.6966} & 0.6827 & 0.6691 & 0.6610 & \best{0.7910} \\
&
& BLEU-1
& 0.3279 & 0.2408 & 0.5441 & 0.3422 & 0.3279 & 0.4792 & 0.5629 & 0.5534
& 0.2999 & \runnerup{0.6432} & 0.6286 & 0.6200 & 0.6069 & \best{0.7479} \\
&
& EM
& 0.2519 & 0.1835 & 0.4161 & 0.2613 & 0.2507 & 0.3402 & 0.4296 & 0.4278
& 0.2279 & \runnerup{0.4985} & 0.4927 & 0.4944 & 0.4757 & \best{0.6347} \\
&
& LLM-Judge
& 0.4331 & 0.3369 & 0.7241 & 0.4643 & 0.4307 & 0.7306 & 0.7431 & 0.7613
& 0.4129 & \runnerup{0.8229} & 0.8016 & 0.7861 & 0.7797 & \best{0.8735} \\
\bottomrule
\end{tabular}%
}
\label{tab:memgallery_full_comparison}
\end{table*}

Table~\ref{tab:memgallery_3b_comparison} repeats the breakdown on the 3B backbone. The aggregate ordering is unchanged---70.17 F1 against 66.79 for the strongest baseline---but the per-task picture is flatter. The two categories whose answer depends on a state that has to be maintained stay well ahead: conflict detection reaches 67.90 against a baseline band of 29.63--35.80, and temporal reasoning 76.41 against 60.36. Knowledge resolution and visual-centric reasoning keep smaller margins (7.1 and 5.1 points), test-time learning ends in a near-tie with MuRAG (79.69 against 79.66), and factual retrieval, visual-centric search, multi-entity reasoning and refusal sit a few points below the strongest baseline. The difference to the 7B table is therefore not a uniform loss of accuracy: what the smaller backbone preserves is keeping and ordering a state, whereas identifying the current observation and its associated entity is more vulnerable to the loss of model capacity in fine-grained visual recall.

\begin{table*}[!t]
\centering
\scriptsize
\setlength{\tabcolsep}{2.5pt}
\renewcommand{\arraystretch}{1.02}
\caption{Per-task comparison on Mem-Gallery with the Qwen-2.5-VL-3B backbone.}
\resizebox{\textwidth}{!}{%
\begin{tabular}{@{}ccc*{8}{r}!{\vrule width 0.5pt}*{6}{r}@{}}
\toprule
\rowcolor{HeaderLavender}
\multicolumn{3}{c}{\textbf{Qwen-2.5-VL-3B}}
& FC (T)
& FIFO
& Naive RAG
& Gen. Agent
& Reflexion
& MemGPT
& A-Mem
& MemoryOS
& FC (V)
& MuRAG
& Universal RAG
& NGM
& AUGUSTUS
& \textbf{LMMR (Ours)} \\
\midrule
\rowcolor{white}

\multirow{12}{*}{\rotatebox[origin=c]{90}{\scriptsize Extract.\ \& Adapt.}}
& \multirow{4}{*}{FR}
& F1
& 0.2501 & 0.1578 & 0.5664 & 0.2301 & 0.2467 & 0.4171 & 0.5985 & 0.5889
& 0.2282 & \runnerup{0.6504} & \best{0.6601} & 0.6045 & 0.5660 & 0.6268 \\
&
& BLEU-1
& 0.2000 & 0.1138 & 0.4862 & 0.1847 & 0.1976 & 0.3494 & 0.4937 & 0.4897
& 0.1795 & \runnerup{0.5673} & \best{0.5811} & 0.5290 & 0.4868 & 0.5283 \\
&
& EM
& 0.0959 & 0.0502 & 0.2740 & 0.0868 & 0.0959 & 0.1872 & 0.2843 & 0.3014
& 0.0868 & \best{0.3699} & \best{0.3699} & \runnerup{0.3470} & 0.3105 & 0.3288 \\
&
& LLM-Judge
& 0.3105 & 0.1918 & 0.7443 & 0.2922 & 0.3082 & 0.5845 & 0.7690 & 0.7694
& 0.2717 & \best{0.8584} & \runnerup{0.8516} & 0.7694 & 0.7626 & 0.7945 \\
\cmidrule(lr){2-17}

& \multirow{4}{*}{VS}
& F1
& 0.2687 & 0.1175 & 0.7424 & 0.3244 & 0.2691 & 0.2631 & 0.7653 & 0.7771
& 0.2154 & \best{0.8661} & 0.8395 & 0.8493 & \runnerup{0.8537} & 0.8113 \\
&
& BLEU-1
& 0.2466 & 0.1012 & 0.6934 & 0.2826 & 0.2471 & 0.2381 & 0.7096 & 0.7123
& 0.1959 & \best{0.8337} & 0.8056 & \runnerup{0.8066} & 0.8047 & 0.7626 \\
&
& EM
& 0.1569 & 0.0359 & 0.5033 & 0.1634 & 0.1601 & 0.1176 & 0.5802 & 0.5458
& 0.0980 & \runnerup{0.6307} & 0.6275 & \best{0.6569} & 0.6242 & 0.6144 \\
&
& LLM-Judge
& 0.2337 & 0.0801 & 0.7092 & 0.2696 & 0.2337 & 0.2843 & 0.6794 & 0.7565
& 0.1797 & \best{0.8676} & 0.8350 & 0.8284 & \runnerup{0.8464} & 0.7827 \\
\cmidrule(lr){2-17}

& \multirow{4}{*}{TTL}
& F1
& 0.4657 & 0.3640 & 0.6769 & 0.5454 & 0.4633 & 0.4833 & 0.6448 & 0.6595
& 0.4857 & \runnerup{0.7966} & 0.7227 & 0.7732 & 0.7328 & \best{0.7969} \\
&
& BLEU-1
& 0.4279 & 0.3161 & 0.6253 & 0.4845 & 0.4251 & 0.4427 & 0.5884 & 0.5997
& 0.4397 & \runnerup{0.7402} & 0.6655 & 0.7145 & 0.6803 & \best{0.7668} \\
&
& EM
& 0.3294 & 0.2344 & 0.5104 & 0.3769 & 0.3234 & 0.3620 & 0.4927 & 0.4777
& 0.3353 & \runnerup{0.6261} & 0.5668 & 0.6083 & 0.5786 & \best{0.6320} \\
&
& LLM-Judge
& 0.6365 & 0.5415 & 0.7433 & 0.6706 & 0.6380 & 0.6276 & 0.7080 & 0.7567
& 0.6320 & \runnerup{0.8338} & 0.7982 & 0.8294 & 0.8027 & \best{0.8546} \\
\cmidrule(lr){2-17}

\multirow{12}{*}{\rotatebox[origin=c]{90}{\scriptsize Reasoning}}
& \multirow{4}{*}{TR}
& F1
& 0.2866 & 0.2398 & 0.4785 & 0.3288 & 0.2866 & 0.3996 & \runnerup{0.6036} & 0.4942
& 0.2972 & 0.5700 & 0.5484 & 0.5332 & 0.5422 & \best{0.7641} \\
&
& BLEU-1
& 0.2688 & 0.2226 & 0.4454 & 0.3036 & 0.2683 & 0.3728 & \runnerup{0.5817} & 0.4654
& 0.2807 & 0.5369 & 0.5119 & 0.4986 & 0.5072 & \best{0.7454} \\
&
& EM
& 0.1301 & 0.0976 & 0.3089 & 0.1626 & 0.1301 & 0.2114 & \runnerup{0.4569} & 0.3089
& 0.1382 & 0.3984 & 0.3821 & 0.3659 & 0.3740 & \best{0.6179} \\
&
& LLM-Judge
& 0.3252 & 0.2480 & 0.6667 & 0.3821 & 0.3211 & 0.5813 & 0.6940 & 0.6707
& 0.3130 & 0.7602 & \runnerup{0.7886} & 0.6951 & 0.7561 & \best{0.8374} \\
\cmidrule(lr){2-17}

& \multirow{4}{*}{VR}
& F1
& 0.2545 & 0.1532 & 0.2966 & 0.1953 & 0.2552 & 0.3226 & 0.3441 & 0.3860
& 0.2285 & 0.4440 & 0.4554 & 0.4260 & \runnerup{0.4681} & \best{0.5194} \\
&
& BLEU-1
& 0.2398 & 0.1384 & 0.2778 & 0.1823 & 0.2395 & 0.3065 & 0.3319 & 0.3685
& 0.2176 & 0.4230 & 0.4352 & 0.4100 & \runnerup{0.4504} & \best{0.4971} \\
&
& EM
& 0.2011 & 0.1092 & 0.1897 & 0.1379 & 0.2011 & 0.2414 & 0.2806 & 0.2931
& 0.1839 & 0.3333 & 0.3333 & 0.3333 & \runnerup{0.3563} & \best{0.4368} \\
&
& LLM-Judge
& 0.3161 & 0.1810 & 0.3822 & 0.2615 & 0.3190 & 0.4310 & 0.4101 & 0.5172
& 0.2615 & 0.5374 & \runnerup{0.5489} & 0.5201 & \best{0.5690} & \best{0.5690} \\
\cmidrule(lr){2-17}

& \multirow{4}{*}{MR}
& F1
& 0.2600 & 0.1934 & 0.4380 & 0.2683 & 0.2615 & 0.3400 & 0.4401 & 0.4483
& 0.2400 & \runnerup{0.4851} & \best{0.4908} & 0.4697 & 0.4773 & 0.4482 \\
&
& BLEU-1
& 0.1943 & 0.1364 & 0.3277 & 0.1997 & 0.1969 & 0.2599 & 0.3162 & 0.3474
& 0.1787 & \runnerup{0.3736} & \best{0.3788} & 0.3602 & 0.3639 & 0.3227 \\
&
& EM
& 0.0291 & 0.0243 & 0.0825 & 0.0388 & 0.0291 & 0.0340 & 0.0722 & \runnerup{0.0971}
& 0.0194 & 0.0874 & 0.0922 & 0.0874 & \best{0.1019} & 0.0777 \\
&
& LLM-Judge
& 0.4442 & 0.2985 & 0.7670 & 0.4345 & 0.4442 & 0.6092 & 0.7474 & 0.7597
& 0.4223 & \runnerup{0.8083} & \best{0.8301} & 0.7985 & 0.7840 & 0.7063 \\
\cmidrule(lr){2-17}

\multirow{12}{*}{\rotatebox[origin=c]{90}{\scriptsize Knowledge Management}}
& \multirow{4}{*}{CD}
& F1
& 0.3333 & \runnerup{0.3580} & 0.3210 & 0.3210 & 0.3333 & \runnerup{0.3580} & 0.3200 & 0.2963
& 0.3457 & 0.3086 & 0.3086 & 0.3086 & 0.3333 & \best{0.6790} \\
&
& BLEU-1
& 0.3333 & \runnerup{0.3580} & 0.3210 & 0.3210 & 0.3333 & \runnerup{0.3580} & 0.3200 & 0.2963
& 0.3457 & 0.3086 & 0.3086 & 0.3086 & 0.3333 & \best{0.6790} \\
&
& EM
& 0.3333 & \runnerup{0.3580} & 0.3210 & 0.3210 & 0.3333 & \runnerup{0.3580} & 0.3200 & 0.2963
& 0.3457 & 0.3086 & 0.3086 & 0.3086 & 0.3333 & \best{0.6790} \\
&
& LLM-Judge
& 0.3333 & \runnerup{0.3580} & 0.3210 & 0.3210 & 0.3333 & \runnerup{0.3580} & 0.3200 & 0.2963
& 0.3457 & 0.3086 & 0.3086 & 0.3086 & 0.3333 & \best{0.6914} \\
\cmidrule(lr){2-17}

& \multirow{4}{*}{KR}
& F1
& 0.2034 & 0.1563 & 0.3382 & 0.2613 & 0.2061 & 0.2372 & \runnerup{0.4280} & 0.3706
& 0.2007 & 0.3975 & 0.3883 & 0.3441 & 0.3604 & \best{0.4988} \\
&
& BLEU-1
& 0.1627 & 0.1118 & 0.2754 & 0.2085 & 0.1639 & 0.1869 & \runnerup{0.3693} & 0.3094
& 0.1625 & 0.3358 & 0.3323 & 0.2842 & 0.2987 & \best{0.4363} \\
&
& EM
& 0.0741 & 0.0370 & 0.1111 & 0.0864 & 0.0741 & 0.0741 & \runnerup{0.2267} & 0.1235
& 0.0864 & 0.1481 & 0.1728 & 0.1358 & 0.1235 & \best{0.2593} \\
&
& LLM-Judge
& 0.4012 & 0.2963 & 0.5926 & 0.4012 & 0.3951 & 0.5185 & 0.6600 & \runnerup{0.6852}
& 0.3704 & 0.6790 & \runnerup{0.6852} & 0.6420 & \best{0.6914} & 0.6667 \\
\cmidrule(lr){2-17}

& \multirow{4}{*}{AR}
& F1
& \runnerup{0.9742} & \best{0.9783} & 0.8942 & 0.9578 & \runnerup{0.9742} & 0.8387 & 0.9080 & 0.8992
& 0.9647 & 0.8823 & 0.8661 & 0.8707 & 0.8553 & 0.9480 \\
&
& BLEU-1
& \runnerup{0.9730} & \best{0.9783} & 0.8924 & 0.9571 & \runnerup{0.9730} & 0.8355 & 0.9080 & 0.8982
& 0.9628 & 0.8811 & 0.8648 & 0.8702 & 0.8540 & 0.9420 \\
&
& EM
& \runnerup{0.9728} & \best{0.9783} & 0.8913 & 0.9565 & \runnerup{0.9728} & 0.8315 & 0.9080 & 0.8967
& 0.9620 & 0.8804 & 0.8641 & 0.8696 & 0.8533 & 0.9130 \\
&
& LLM-Judge
& \best{0.9620} & \best{0.9620} & 0.8804 & 0.9429 & \best{0.9620} & 0.8234 & 0.9052 & 0.8804
& \runnerup{0.9484} & 0.8641 & 0.8478 & 0.8587 & 0.8424 & 0.9130 \\
\cmidrule(lr){2-17}

\multirow{4}{*}{}
& \multirow{4}{*}{Overall}
& F1
& 0.3798 & 0.2986 & 0.5832 & 0.4013 & 0.3793 & 0.4165 & 0.6059 & 0.6013
& 0.3665 & \runnerup{0.6679} & 0.6479 & 0.6443 & 0.6383 & \best{0.7017} \\
&
& BLEU-1
& 0.3492 & 0.2689 & 0.5333 & 0.3620 & 0.3487 & 0.3794 & 0.5505 & 0.5462
& 0.3360 & \runnerup{0.6194} & 0.5995 & 0.5953 & 0.5881 & \best{0.6521} \\
&
& EM
& 0.2624 & 0.2040 & 0.3933 & 0.2671 & 0.2618 & 0.2700 & 0.4303 & 0.4103
& 0.2496 & \runnerup{0.4728} & 0.4594 & 0.4670 & 0.4541 & \best{0.5172} \\
&
& LLM-Judge
& 0.4541 & 0.3521 & 0.6856 & 0.4597 & 0.4538 & 0.5383 & 0.6886 & 0.7162
& 0.4272 & \best{0.7756} & \runnerup{0.7662} & 0.7463 & 0.7528 & \best{0.7756} \\
\bottomrule
\end{tabular}%
}
\label{tab:memgallery_3b_comparison}
\end{table*}

\subsection{Per-setting results on MemEye}
\label{app:memeye-full}

Table~\ref{tab:memeye_full_comparison} reports MemEye at the resolution the benchmark itself uses: the twelve $(X,Y)$ cells that the main figures aggregate are laid out as rows, each with its multiple-choice EM, its open-ended BLEU-1 and its open-ended judge score, and the columns are the twelve published systems---split into textual and multimodal memory---together with LMMR, which was never trained on a MemEye question. On the unweighted average over the twelve cells LMMR leads all three metrics (EM 0.6696, BLEU-1 0.2425, judge 0.4133, against 0.5361, 0.2171 and 0.4017 for the strongest caption-based baseline, FC (T)), and the per-cell rows show where that lead comes from and where it does not. LMMR takes eleven of the twelve cells on multiple-choice EM, and eight of the twelve on the judge; the cells it does not take are the shallow ones, where a dense caption already carries the decisive detail---$X2/Y1$ (0.3100 against 0.6111) and $X2/Y2$ (0.2178 against 0.3590) are the two clearest reversals. The gains have the opposite shape: the largest single margins appear where fine visual evidence has to survive an update, at $X2/Y3$ (0.9360 against 0.8000) and $X4/Y3$ (0.3215 against 0.1667). The pattern therefore repeats the main-text observation at cell level: captions are competitive when the answer is a region description, and lose when the answer is a pixel or an instance that has changed.

\begin{table*}[t]
\centering
\scriptsize
\setlength{\tabcolsep}{2.8pt}
\renewcommand{\arraystretch}{1.04}
\caption{Per-setting results on MemEye with the Qwen3-VL-8B backbone.}
\resizebox{\textwidth}{!}{%
\begin{tabular}{@{}lll *{7}{r} *{5}{r} r@{}}
\toprule
\multicolumn{3}{c}{} & \multicolumn{7}{c}{\textbf{Textual memory}} & \multicolumn{5}{c}{\textbf{Multimodal memory}} & \\
\cmidrule(lr){4-10} \cmidrule(lr){11-15}
\rowcolor{HeaderLavender}
\textbf{Y} & \textbf{X} & \textbf{Metric} & FC(T) & SRAG(T) & Ref. & Gen.Ag. & MemOS & A-Mem & SM(T) & FC(V) & SRAG(V) & MMA & M2A & SM(V) & \textbf{LMMR (Ours)} \\
\midrule
\multirow{12}{*}{Y1} & \multirow{3}{*}{X1} & EM & 0.8500 & 0.8500 & \runnerup{0.9250} & 0.2500 & 0.7750 & 0.8000 & 0.5250
& 0.5750 & 0.5000 & 0.5500 & 0.5250 & 0.5500 & \best{0.9583} \\
 &  & BLEU-1 & \runnerup{0.2658} & 0.1958 & 0.1976 & 0.1131 & 0.2471 & 0.1537 & 0.2425
& 0.1177 & 0.1093 & 0.1022 & 0.0874 & \best{0.2925} & 0.2178 \\
 &  & LLM-Judge & 0.4500 & \best{0.5500} & 0.4500 & 0.2500 & \best{0.5500} & 0.4000 & \runnerup{0.5000}
& 0.3000 & 0.3500 & 0.1500 & 0.1500 & 0.4500 & 0.4860 \\
\cmidrule(lr){3-16}
 & \multirow{3}{*}{X2} & EM & \best{0.7500} & 0.5278 & 0.6389 & 0.2500 & 0.5278 & 0.5833 & 0.6111
& 0.3056 & 0.4444 & 0.2222 & 0.5278 & 0.6111 & \runnerup{0.7479} \\
 &  & BLEU-1 & 0.1561 & 0.1123 & 0.1471 & 0.0736 & 0.0874 & \runnerup{0.2119} & 0.0150
& 0.1544 & 0.1123 & 0.0069 & 0.0712 & 0.0150 & \best{0.2596} \\
 &  & LLM-Judge & \best{0.6111} & \runnerup{0.3333} & \runnerup{0.3333} & 0.2778 & 0.1111 & \best{0.6111} & 0.2222
& 0.2222 & 0.2778 & 0.0000 & 0.1111 & 0.2222 & 0.3100 \\
\cmidrule(lr){3-16}
 & \multirow{3}{*}{X3} & EM & \runnerup{0.4901} & \runnerup{0.4901} & 0.4605 & 0.2599 & 0.4243 & 0.4507 & 0.1118
& 0.3026 & 0.3586 & 0.3783 & 0.3717 & 0.1118 & \best{0.7004} \\
 &  & BLEU-1 & \runnerup{0.2823} & 0.2570 & 0.2737 & 0.1893 & 0.1690 & 0.1753 & 0.0589
& 0.1792 & 0.2719 & 0.2050 & 0.1192 & 0.0688 & \best{0.3124} \\
 &  & LLM-Judge & 0.3092 & 0.3158 & \runnerup{0.3487} & 0.2303 & 0.2171 & 0.2632 & 0.1053
& 0.2500 & 0.2632 & 0.2237 & 0.2039 & 0.1118 & \best{0.4411} \\
\cmidrule(lr){3-16}
 & \multirow{3}{*}{X4} & EM & 0.5278 & \runnerup{0.6111} & 0.5278 & 0.2500 & 0.5556 & 0.5833 & 0.2639
& 0.4167 & 0.4306 & 0.2083 & 0.2917 & 0.2639 & \best{0.6981} \\
 &  & BLEU-1 & \runnerup{0.2051} & \best{0.2097} & 0.1046 & 0.0273 & 0.1525 & 0.1721 & 0.0556
& 0.0779 & 0.1771 & 0.1378 & 0.0377 & 0.0556 & 0.1603 \\
 &  & LLM-Judge & \best{0.3889} & \runnerup{0.3611} & 0.2778 & 0.0833 & 0.2222 & 0.2778 & 0.1111
& 0.3333 & 0.3056 & 0.2222 & 0.0833 & 0.1111 & 0.3480 \\
\midrule
\multirow{12}{*}{Y2} & \multirow{3}{*}{X1} & EM & 0.4138 & 0.4655 & 0.4310 & 0.2414 & 0.3966 & 0.4052 & 0.3621
& 0.2241 & 0.4397 & \runnerup{0.4741} & 0.3103 & 0.3707 & \best{0.5969} \\
 &  & BLEU-1 & 0.2811 & 0.3167 & 0.3317 & 0.1716 & 0.1968 & 0.2836 & 0.2634
& 0.1179 & \best{0.4219} & 0.2108 & 0.0737 & 0.1939 & \runnerup{0.3781} \\
 &  & LLM-Judge & 0.3621 & 0.3621 & 0.4310 & 0.2414 & 0.2414 & 0.3793 & 0.3276
& 0.2069 & \best{0.4828} & 0.2931 & 0.1034 & 0.2414 & \runnerup{0.4647} \\
\cmidrule(lr){3-16}
 & \multirow{3}{*}{X2} & EM & \runnerup{0.4679} & 0.3397 & 0.4359 & 0.2564 & 0.3654 & 0.4038 & 0.3526
& 0.1923 & 0.4103 & 0.4167 & 0.4295 & 0.3397 & \best{0.4810} \\
 &  & BLEU-1 & \runnerup{0.2243} & 0.1091 & \best{0.2341} & 0.2144 & 0.0576 & 0.0785 & 0.0807
& 0.1542 & 0.1995 & 0.1163 & 0.0909 & 0.0765 & 0.0618 \\
 &  & LLM-Judge & \best{0.3590} & 0.2179 & \runnerup{0.3205} & 0.2692 & 0.1538 & 0.1410 & 0.1538
& 0.1026 & 0.1795 & 0.1154 & 0.0513 & 0.1410 & 0.2178 \\
\cmidrule(lr){3-16}
 & \multirow{3}{*}{X3} & EM & 0.3690 & 0.3452 & 0.3095 & 0.2500 & 0.3393 & 0.3155 & 0.1964
& 0.2024 & 0.2976 & 0.2679 & \runnerup{0.3750} & 0.2262 & \best{0.5005} \\
 &  & BLEU-1 & 0.1373 & 0.1174 & 0.0967 & 0.1155 & \runnerup{0.1423} & 0.1303 & 0.0432
& 0.1012 & 0.1376 & 0.0303 & 0.0933 & 0.0457 & \best{0.1479} \\
 &  & LLM-Judge & \runnerup{0.2500} & 0.2143 & 0.1548 & 0.1310 & 0.2381 & 0.1786 & 0.0952
& 0.1310 & 0.2024 & 0.0000 & 0.2143 & 0.1310 & \best{0.3010} \\
\cmidrule(lr){3-16}
 & \multirow{3}{*}{X4} & EM & 0.3807 & 0.3693 & 0.3494 & 0.2500 & 0.3892 & \runnerup{0.4034} & 0.1506
& 0.2756 & 0.3352 & 0.2869 & 0.3040 & 0.1307 & \best{0.4455} \\
 &  & BLEU-1 & \runnerup{0.2880} & 0.2265 & 0.2654 & 0.2712 & 0.2153 & 0.1769 & 0.0795
& 0.2097 & 0.1744 & 0.1898 & 0.0476 & 0.1003 & \best{0.3083} \\
 &  & LLM-Judge & \runnerup{0.3239} & 0.2443 & 0.3011 & 0.2955 & 0.2216 & 0.2102 & 0.1023
& 0.2557 & 0.2102 & 0.2159 & 0.1250 & 0.1364 & \best{0.3345} \\
\midrule
\multirow{12}{*}{Y3} & \multirow{3}{*}{X1} & EM & 0.5750 & 0.5250 & 0.5250 & 0.2500 & \runnerup{0.7500} & 0.7250 & 0.5250
& 0.2750 & 0.2250 & 0.2750 & 0.4250 & 0.4750 & \best{0.7917} \\
 &  & BLEU-1 & 0.1623 & 0.1427 & 0.1667 & 0.0836 & 0.1283 & 0.2122 & \best{0.2515}
& 0.0805 & 0.0408 & 0.0264 & 0.0428 & \runnerup{0.2413} & 0.1663 \\
 &  & LLM-Judge & 0.5000 & 0.5000 & 0.5000 & 0.4000 & \best{0.5500} & 0.4500 & \best{0.5500}
& 0.2500 & 0.1000 & 0.0000 & 0.0000 & \best{0.5500} & \runnerup{0.5145} \\
\cmidrule(lr){3-16}
 & \multirow{3}{*}{X2} & EM & 0.8000 & 0.7250 & \runnerup{0.8250} & 0.2250 & 0.7000 & 0.6000 & 0.7000
& 0.3500 & 0.5000 & 0.2500 & 0.2500 & 0.7250 & \best{0.8500} \\
 &  & BLEU-1 & 0.2543 & 0.2227 & 0.2147 & 0.1895 & 0.2091 & \runnerup{0.3594} & 0.2910
& 0.2249 & 0.2182 & 0.0367 & 0.0843 & 0.2939 & \best{0.5107} \\
 &  & LLM-Judge & \runnerup{0.8000} & 0.6000 & \runnerup{0.8000} & 0.5500 & 0.4000 & 0.5500 & 0.6000
& 0.4000 & 0.2500 & 0.0000 & 0.0000 & 0.6000 & \best{0.9360} \\
\cmidrule(lr){3-16}
 & \multirow{3}{*}{X3} & EM & \runnerup{0.5500} & 0.5250 & 0.4750 & 0.2500 & 0.4000 & 0.4000 & 0.3000
& 0.3250 & 0.4500 & 0.3750 & 0.4750 & 0.2750 & \best{0.8600} \\
 &  & BLEU-1 & 0.1523 & 0.0956 & \best{0.1541} & 0.0979 & 0.1053 & \runnerup{0.1540} & 0.0111
& 0.0764 & 0.1096 & 0.1086 & 0.1054 & 0.0091 & 0.1316 \\
 &  & LLM-Judge & \best{0.3000} & 0.1000 & 0.2500 & 0.0000 & \best{0.3000} & 0.2500 & 0.0000
& 0.0500 & 0.2000 & 0.0500 & 0.0500 & 0.0000 & \runnerup{0.2844} \\
\cmidrule(lr){3-16}
 & \multirow{3}{*}{X4} & EM & 0.2583 & 0.1667 & 0.2083 & 0.2500 & 0.2250 & 0.1917 & 0.0250
& 0.1250 & 0.1833 & 0.2500 & \runnerup{0.3417} & 0.0417 & \best{0.4051} \\
 &  & BLEU-1 & 0.1967 & 0.1245 & \runnerup{0.2085} & 0.1504 & 0.0603 & 0.0896 & 0.1191
& 0.0048 & 0.1757 & 0.0770 & 0.0413 & 0.1191 & \best{0.2556} \\
 &  & LLM-Judge & 0.1667 & 0.1333 & 0.2000 & \runnerup{0.2167} & 0.0833 & 0.1167 & 0.1000
& 0.0333 & 0.2000 & 0.0667 & 0.1333 & 0.1000 & \best{0.3215} \\
\midrule
\multirow{3}{*}{Avg.} &  & EM & \runnerup{0.5361} & 0.4950 & 0.5093 & 0.2486 & 0.4873 & 0.4885 & 0.3436
& 0.2974 & 0.3812 & 0.3295 & 0.3856 & 0.3434 & \best{0.6696} \\
 &  & BLEU-1 & \runnerup{0.2171} & 0.1775 & 0.1996 & 0.1415 & 0.1476 & 0.1831 & 0.1260
& 0.1249 & 0.1790 & 0.1040 & 0.0746 & 0.1260 & \best{0.2425} \\
 &  & LLM-Judge & \runnerup{0.4017} & 0.3277 & 0.3639 & 0.2454 & 0.2741 & 0.3190 & 0.2390
& 0.2112 & 0.2518 & 0.1114 & 0.1021 & 0.2329 & \best{0.4133} \\
\bottomrule
\end{tabular}%
}
\label{tab:memeye_full_comparison}
\end{table*}

\section{Additional analyses}
\label{app:analyses}

\subsection{Cross-modal alignment of the shared concept space}
\label{app:alignment}

We measure caption--image retrieval using Top-1 accuracy and mean reciprocal rank (MRR).

\begin{figure}[!t]
  \centering
  \includegraphics[width=0.9\textwidth]{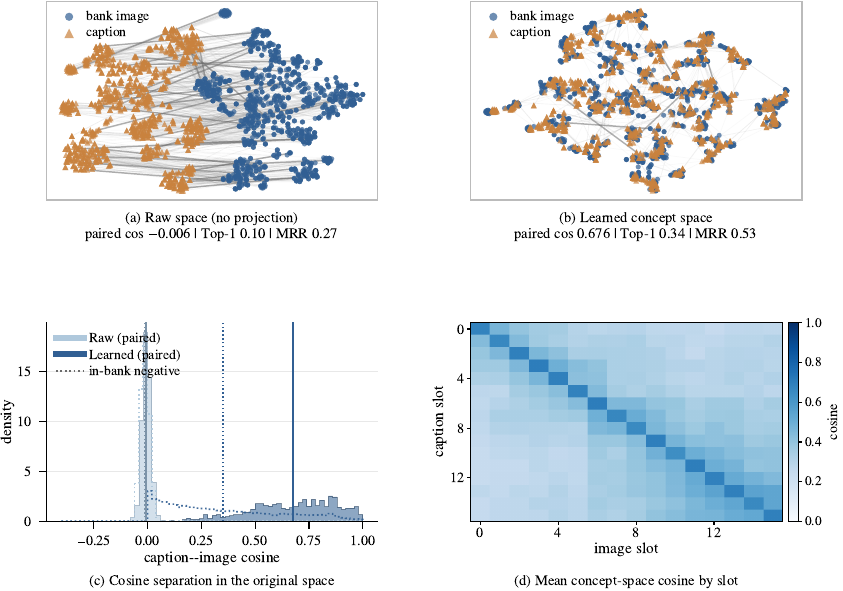}
  \caption{\textbf{Cross-modal alignment of the shared concept space} (Qwen2.5-VL-7B).
  Each point is a caption (triangles) or a bank image (circles), with matched pairs joined by lines:
  (a)~before any cross-modal projection the two modalities form separate clouds, whereas (b)~in the trained concept space they mix and the pair lines shorten.
  (c)~The same effect measured in the original space, where solid histograms are paired captions and images and dotted ones are the same-bank negatives: the alignment objective moves the paired cosine from $-0.006$ to $0.676$ and within-bank Top-1 from $0.10$ to $0.34$, against a chance level of $0.07$.
  (d)~Mean cosine between caption slot $i$ and image slot $j$: the diagonal reaches $0.68$ while the off-diagonal cells stay at $0.35$, so what the objective produces is a separable slot structure rather than a collapse of the space.
  The map is qualitative and the pairs are caption--own-image pairs, so the figure audits cross-modal alignment rather than the full retrieval and answer-generation pipeline used in the downstream memory task.}
  \label{fig:modality-tsne}
\end{figure}

Fig.~\ref{fig:modality-tsne} checks the premise the rest of the method rests on: that the shared dictionary really does give the two modalities a common address space, rather than two coordinate systems that happen to be trained at the same time. We compare two representation spaces on exactly the same $3{,}429$ caption--image pairs, taken from $240$ banks of the evaluation questions (mean candidate set $14.3$). The raw space is the untrained reference: the language model's mean-pooled caption hidden state against the mean-pooled visual feature, with no cross-modal projection at all. The learned space is the one LMMR uses: both sides are pushed through their projectors into the dictionary, and the alignment objective couples the two modalities within this shared representation space.

Three things follow from the comparison, and all three are visible in the figure. First, in the raw space the two modalities are unrelated: the paired cosine is $-0.006$, i.e.\ slightly \emph{below} the $-0.012$ of an in-bank negative, and retrieval within a bank is $10.1\%$ against a chance rate of $7.0\%$. Two clouds, one per modality, with the pair lines crossing the whole map. Second, the alignment objective closes that gap without collapsing the space: the paired cosine rises to $0.676$ while in-bank negatives sit at $0.351$, a margin of $0.32$; mean rank drops from $6.90$ to $3.40$ and MRR rises from $0.27$ to $0.53$. Panel~(d) makes the same point slot by slot, in the matrix whose $(i,j)$ entry is the mean cosine between caption slot $i$ and image slot $j$ across banks: the diagonal averages $0.68$ and the off-diagonal $0.35$, so the objective separates the paired slots instead of pulling every slot into one direction. Third, on an absolute scale the picture stays honest: $34.4\%$ Top-1 within a bank is roughly five times chance, but it is not near-ceiling, which is what one should expect from a dictionary that is trained jointly with the generator rather than on a retrieval objective alone. The panel is therefore an argument about \emph{addressability}---a textual query and the image it refers to end up in the same neighbourhood of the concept space---and not a claim that retrieval through the dictionary is already solved.

\subsection{A quantitative alignment probe}
\label{app:probe}

\begin{figure}[!t]
  \centering
  \includegraphics[width=0.9\textwidth]{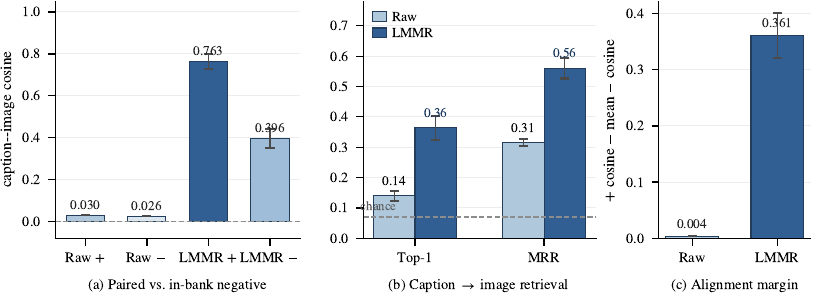}
  \caption{\textbf{Cross-modal alignment probe} (Qwen3-VL-8B reader).
  A caption is paired with its own bank image (positive) and with the remaining images of the same bank (negatives; mean candidate set $14.3$).
  (a)~Paired and in-bank negative caption--image cosines before and after training: the raw representations do not separate the two, whereas in the shared concept space the paired cosine reaches $0.76$ against $0.40$ for the negatives.
  (b)~Caption$\rightarrow$image retrieval within a bank (Top-1 and MRR), with the chance level of Top-1 dashed.
  (c)~The margin of (a) as a single number.
  Bars are means and error bars are one standard deviation over the evaluation splits.}
  \label{fig:alignment-probe}
\end{figure}

Fig.~\ref{fig:alignment-probe} measures the same property as Fig.~\ref{fig:modality-tsne} in two different ways: it uses the reader that answers the MemEye questions rather than the Mem-Gallery one, and it reports the pairing statistics directly instead of through an embedding. Over $4{,}835$ caption--image pairs on average per evaluation split, the raw representations are again unusable as an address: the paired cosine is $0.030$ against $0.026$ for the in-bank negatives, a margin of $0.004$, and retrieval reaches $14.0\%$ Top-1. The shared concept space separates the two populations by two orders of magnitude in margin: $0.763$ paired against $0.396$ negative (margin $0.361$), with Top-1 rising to $36.4\%$ and MRR to $0.560$ from $0.315$. The two panels therefore agree across reader families and across measurement styles, and the ordering is what the selection stage needs: on average a caption sits closer to its own image than to the other images of the same bank, so a textual query can address visual evidence while retaining the image representation instead of replacing its visual content with a textual summary.

\section{Prompts}
\label{app:prompts}

\providecolor{PromptFill}{RGB}{238, 234, 248}
\providecolor{PromptRule}{RGB}{150, 140, 185}
\lstdefinestyle{prompt}{
  basicstyle=\ttfamily\footnotesize,
  backgroundcolor=\color{PromptFill},
  rulecolor=\color{PromptRule},
  frame=single, framerule=0.4pt,
  framesep=4pt, framexleftmargin=4pt, framexrightmargin=4pt,
  xleftmargin=6pt, xrightmargin=6pt,
  breaklines=true, breakatwhitespace=false, columns=fullflexible, keepspaces=true,
  aboveskip=6pt, belowskip=6pt, showstringspaces=false,
}

This appendix reproduces the prompts used for the reported numbers. The answering prompts follow the official templates of the two benchmarks; the judge prompts are the ones that produced the scores of Tables~\ref{tab:memgallery_overall_size_comparison} and~\ref{tab:memeye_full_comparison}. Placeholders are shown in braces.

\subsection{Answer generation}
\label{app:prompt-answer}

\noindent\textbf{System prompt (Mem-Gallery, official).}
\begin{lstlisting}[style=prompt]
You are an AI assistant evaluated on multimodal long-term conversational memory.
For the given question-answering task, your responses must be concise, yet complete enough to accurately answer the questions.
If multiple pieces of information about the same event appear in the conversation, always rely on the most recent information.

The question-answering evaluation will contain several multimodal task types:

Factual Retrieval: Retrieve explicit facts mentioned in the conversation for the answer.
Multi-entity Reasoning: Combine the retrieved information to reason and infer an answer.
Temporal Reasoning: Resolve time-dependent questions.
Visual-centric Reasoning: Besides textual information, answer questions using visual images in the conversation.
Test-time Learning: Learn new visual knowledge from provided images within historical dialogue and use it in question-answering.
Visual-centric Search: Find the image(s) that match the information in a given query and return their image ID(s).
Conflict Detection: Detect contradictions between the conversation history and the information provided in the question.
Knowledge Resolution: Resolve knowledge conflicts or updates by prioritizing the most recent information.
Answer Refusal: Decline to answer when the information does not exist in the conversation history.

Follow all instructions strictly. Only answer using information contained within the multimodal conversation. Do not hallucinate. Always remain consistent and grounded in the dialogue history.
\end{lstlisting}

\noindent\textbf{User turn (memory block, latent hint, question).} The memory block carries the retrieved context; the latent hint is our only addition, and it asks the model to use the latent rollout while keeping intermediate reasoning in the continuous state rather than expressing it as generated text.
\begin{lstlisting}[style=prompt]
The retrieved memory contents are as follows:

{memory_context}

Think silently in your own latent space before answering. When a step feels uncertain, recall the corresponding visual detail from memory, then continue.

Your task is to answer the question about the conversation between the user and the assistant in a concise manner with the help of memory content.
Please only provide the content of the answer, without including introductory phrases like "answer:".
For questions that require answering a date or time, strictly follow the format and provide a specific date or time whenever possible.
Generate answers primarily concise, yet complete enough to accurately answer the questions.

The current question is as follows:
{question} {format_constraint}
\end{lstlisting}

\noindent\textbf{Per-task format constraints} (appended to questions in the three task categories requiring the specific answer formats shown below for consistent parsing and scoring).
\begin{lstlisting}[style=prompt]
AR: Provide your answer based on the information in the conversation. Only if the information about the question is not present in the conversation, reply with: "Not mentioned."
CD: Please check whether this information conflicts with the conversation, and reply strictly with either "Yes." or "No."
VS: Return the image_id of the image(s). If there are multiple images, sort them in ascending order and separate them by commas. Format example: "D2:IMG_003, D2:IMG_010, D10:IMG_002" (for format reference only).
\end{lstlisting}

\noindent\textbf{Question-image caption (MemEye).} When a question carries an image, its caption is generated once with the prompt below and spliced in front of the question text, so that the reader knows what the picture shows without the runtime visual path having to read it out again.
\begin{lstlisting}[style=prompt]
Describe this image faithfully and densely. Include visible text/OCR, people, objects, attributes, spatial relations, and actions. Do not infer facts that are not visible. Return only the description.

The picture in the question shows: {question_image_caption}
Question: {question}
\end{lstlisting}

\subsection{Judge prompts}
\label{app:prompt-judge}

The judges return structured objects containing a score and a brief explanation.

\noindent\textbf{Mem-Gallery (official rubric, scored with Qwen2.5-72B-Instruct).}
\begin{lstlisting}[style=prompt]
You are an impartial judge evaluating the memory capabilities of an AI assistant with the question-answering task.
Your task is to compare the Assistant's Answer against the Ground Truth and assign a score of 0, 0.25, 0.5, 0.75, or 1.

### Scoring Rubric

**Score 0 (Incorrect / Miss):**
- The answer contradicts the Ground Truth.
- For Yes/No questions: The answer has the wrong polarity (e.g., says "Yes" when Ground Truth is "No").
- For Open-ended questions: The answer provides factually wrong information or hallucinations.
- The assistant fails to provide the required information.

**Score 0.25 (Poor / Tangential):**
- The answer touches on the topic but misses the core entity or key value required.
- The answer contains a mix of minor correct details and significant hallucinations or wrong associations.
- The answer is excessively vague to the point of being useless (e.g., answering "a dog" instead of "a golden retriever").

**Score 0.5 (Partial / Vague):**
- The answer is technically correct, but lacks confidence or is incomplete.
- The answer captures the main entity or concept correctly but misses a part of the required supporting details.
- For Yes/No questions: The polarity is correct, but the reasoning is flawed (if have), or the assistant is uncertain (e.g., "I think it might be Yes").
- For Open-ended questions: The answer is too general or misses key adjectives/details present in the Ground Truth.

**Score 0.75 (Good / Minor Imperfection):**
- The answer is largely accurate and captures the core information confidently.
- It misses only minor details (e.g., specific adjectives or secondary details) that do not alter the main truth.
- The answer contains all the correct information but includes unnecessary "fluff" or slight conversational filler that reduces precision.

**Score 1 (Correct / Exact):**
- The answer is accurate, precise, and confident.
- For Yes/No questions: The polarity matches the Ground Truth perfectly.
- For Open-ended questions: The answer contains all the core information and necessary details required by the Ground Truth without hallucinations.

### Input Data

Question: {question}
Ground Truth: {ground_truth}
Assistant Answer: {model_output}

### Output Format

Output strictly in the following JSON format:

{"score": <0, 0.25, 0.5, 0.75, or 1>, "reasoning": "<short explanation>"}
\end{lstlisting}

\noindent\textbf{MemEye (official protocol, scored with GPT-5.2).}
\begin{lstlisting}[style=prompt]
You are an impartial judge evaluating a memory-augmented question answering system. Compare the Assistant's Answer against the Ground Truth answer and assign a normalized semantic-correctness score in [0,1], where 1 means the Assistant's Answer conveys exactly the same information as the Ground Truth answer (minor wording differences are fine), 0.5 means it is partially correct or misses important information, and 0 means it is wrong or contradictory. Ignore differences in phrasing, grammar, and answer format.

Question: {question}
Ground Truth: {ground_truth}
Assistant's Answer: {model_output}

Respond with a JSON object only: {"score": <float in [0,1]>, "rationale": "<one short sentence>"}
\end{lstlisting}

\end{document}